\documentclass[journal,onecolumn,12pt,draftclsnofoot]{IEEEtran}
\usepackage[utf8]{inputenc}
\usepackage[normalem]{ulem}
\usepackage{multirow}
\usepackage{graphicx}
\usepackage{cite}
\usepackage{tikz}
\usepackage[autostyle=true,english=american]{csquotes}

\graphicspath{{./images/}}

\begin{document}

\title{Paraphrase Identification with Deep Learning: A Review of Datasets and Methods}

\author{Chao~Zhou,
        Cheng~Qiu,
        Lizhen~Liang,
        and~Daniel~E.~Acuna\thanks{Corresponding author: daniel.acuna@colorado.edu}%
\thanks{C. Zhou did this work while a master's student at Syracuse University, NY, USA).}%
\thanks{C. Zhou is with SingularDance, Shanghai, 200081 China (e-mail: joseph.zhou@singulardance.com).}%
\thanks{C. Qiu is with the College of Arts and Science, Vanderbilt University, Nashville, TN 37235 USA (e-mail: cheng.qiu@vanderbilt.edu).}%
\thanks{L. Liang is with the School of Information Science, Syracuse University, Syracuse, NY 13244 USA (e-mail: lliang09@syr.edu).}%
\thanks{D. E. Acuna is with the Department of Computer Science, University of Colorado at Boulder, Boulder, CO 80309 USA.}%
}

\maketitle

\begin{abstract}
The rapid progress of Natural Language Processing (NLP) technologies has led to the widespread availability and effectiveness of text generation tools such as ChatGPT and Claude. While highly useful, these technologies also pose significant risks to the credibility of various media forms if they are employed for paraphrased plagiarism---one of the most subtle forms of content misuse in scientific literature and general text media. Although automated methods for paraphrase identification have been developed, detecting this type of plagiarism remains challenging due to the inconsistent nature of the datasets used to train these methods. In this article, we examine traditional and contemporary approaches to paraphrase identification, investigating how the under-representation of certain paraphrase types in popular datasets, including those used to train Large Language Models (LLMs), affects the ability to detect plagiarism. We introduce and validate a new refined typology for paraphrases (\textsc{ReParaphrased}, REfined PARAPHRASE typology definitions) to better understand the disparities in paraphrase type representation. Lastly, we propose new directions for future research and dataset development to enhance AI-based paraphrase detection.
\end{abstract}
\IEEEpeerreviewmaketitle
\section{Introduction}
\IEEEPARstart{T}{he} development of new text generation methods, such as GPT-3 \cite{brown2020language} and ChatGPT \cite{ChatGPT}, has facilitated the creation of paraphrased text. Such advancement raises concerns about the ability to accurately identify paraphrased content in order to protect the credibility of media sources and science. Automated paraphrase identification approaches are being developed and used to detect paraphrased text, but the limitations of current methods have been discovered and discussed. A study that examined fifteen common plagiarism detection tools \cite{foltynek_testing_2020} found that most of these tools were unable to detect paraphrased content. In our paper, we investigate traditional language models and deep learning methods for paraphrase identification, and examine the limitations of current datasets, methods, and evaluation metrics used in this field \cite{li_simple_2016,elder_towards_2018,Holtzman2020The,yang_diversity_2021,yang_improving_2022,tevet_evaluating_2021}. The paper also presents a typology of paraphrases that can be used for constructing new datasets, and proposes potential directions for future research in this area. We also discuss the unprecedented threats posed by Large Language Models (LLMs).

Paraphrase identification (PI) is beneficial for several significant natural language processing (NLP) applications. For instance, a summarization system can eliminate redundancy when generating summaries by removing paraphrases \cite{gupta_abstractive_2019}. Additionally, paraphrases can be used to improve the performance of question-answering systems \cite{clough_developing_2011}. Perhaps most notably, paraphrase identification is essential for detecting plagiarism. In education \cite{nielsen2009recognizing}, paraphrase identification can be used to evaluate whether a student's submission or answer is semantically equivalent to a reference answer, or to detect paraphrased content in academic papers \cite{roe2022automated}.

There is a rich history of datasets and methods for paraphrase identification. Traditional approaches typically rely on hand-crafted rules \cite{mckeown_paraphrasing_1983, hassan-etal-2007-supertagged}, whereas more recent techniques have leveraged deep learning to achieve improved results. Many classic NLP tasks, such as machine translation \cite{bahdanau_neural_2014, cho-etal-2014-learning}, text summarization \cite{nallapati-etal-2016-abstractive}, and dialogue systems \cite{serban-etal-2017-piecewise}, can be viewed as paraphrase identification and generation tasks. As such, advances in these areas have also translated into improvements in paraphrase identification. For example, a number of studies have demonstrated that deep learning models can achieve state-of-the-art performance for detecting sophisticated paraphrases \cite{prakash_neural_2016, gupta_abstractive_2019, li_paraphrase_2018}. The long history of this field is closely tied to the development of artificial intelligence itself.

Although many paraphrase identification datasets and methods have been proposed, a comprehensive understanding of their achievements and challenges is still lacking. The main challenges of paraphrase identification lie in two aspects: First, no large and balanced training datasets exist for paraphrase identification. Most manually labeled training corpus has limited data points because labeling paraphrases requires extensive linguistic knowledge. At the same time, large-size auto-generated datasets are corpus that are translated forward and backward (known as back-translations), limiting paraphrase structures. Second, detection models fail to detect long and complex paraphrased passages with sophisticated modifications. Traditional methods mainly focus on paraphrased sentences' lexical and syntactic information, making it hard to represent and detect complex semantic features. Recent deep learning neural network models can extract and represent semantic information from paraphrased well by learning from the training corpus. However, many of the datasets used for these models lack a well-balanced distribution of paraphrase types, and in some cases, do not include certain types of crucial paraphrases. For example, the same-polarity substitution type, which replaces a synonym with the target word, is one of the most common paraphrase types in back-translation corpora. In contrast, opposite-polarity substitution should modify the structure using antonyms to maintain semantic consistency. This method is commonly found in human literature yet it is empirically hard to see in the generated training corpus (We have experiments showing this empirical result in section V).

In this survey, we explore how the types of paraphrases present in standard datasets impact identification and generation tasks. We provide a comprehensive review of these datasets and summarize the different classification schema of paraphrases that different datasets contain. Previous research has focused on classifying paraphrase identification methods \cite{el2019exploring, XIAO2020172}, but has not examined the crucial relationship between datasets and methods. As such, we also review a range of traditional and modern (e.g. deep learning-based) methods for paraphrase identification. Our results suggest that identification accuracy can be improved by carefully curating the paraphrase type distributions in training datasets. Furthermore, we discuss how these distributions can affect paraphrase generation. Our key contributions include:

\begin{enumerate}
\item A refined paraphrase typology
\item A review of paraphrase identification methods
\item An automatic paraphrase type classifier
\item A new typology of paragraphs and their distribution across datasets
\item A discussion of how to create a dataset that will maximize the paraphrase identification accuracy.
\end{enumerate}

\section{Definition and typology of paraphrasing}
According to the Oxford Language Dictionary, paraphrasing is defined as "a rewording of something written or spoken by someone else" \cite{hornby_oxford_1995}. In other words, paraphrases are sentences or phrases that convey the same underlying semantic meaning using different wording. For the purpose of analysis, this definition lacks the specificity of the properties that constitute semantic equivalence between two texts. For instance, there may be non-paraphrases with similar structures and words but different meanings. Thus, a more concrete definition is necessary to more precisely encapsulate the concept of semantic equivalence.

Past research has also investigated the definition of semantic equivalence from various perspectives. In textual entailment, semantic equivalence is often viewed as a bidirectional relationship between two texts, where the meaning of one text can be inferred from the context of the other, and vice versa \cite{dagan_pascal_2005}. From the perspective of propositional logic, semantic equivalence is defined as the degree of symmetry between two texts, such that one text is a (not necessarily proper) subset of the other \cite{dras_tree_nodate}. This definition allows for paraphrases that do not perfectly capture the meaning of the original text. Alternatively, \cite{harris_distributional_1954} argued that the morphological structure of a sentence is not uniquely linked to its meaning and that the semantic meaning of a sentence is instead determined by the specific elements that make it up. From this perspective, semantic equivalence can be measured by the distribution of words, where sentences with a high degree of word overlap in similar contexts are considered semantically equivalent \cite{lin_dirt_2001}. More recently, \cite{bhagat-hovy-2013-squibs} proposed that paraphrases should exhibit properties of a paraphrase category known as Quasi-paraphrases, which are defined as "approximate equivalence that conveys similar meanings using different words". They identified 25 distinct paraphrase operations that can be used to produce quasi-paraphrases. While this definition is less restrictive than others, it may still be susceptible to ambiguity, as it does not account for the speaker's perspective or evaluation of a situation.

Given the ambiguity in its definition, paraphrases should be defined more concretely than is commonly understood. We adopt the definition of quasi-paraphrases proposed by \cite{bhagat-hovy-2013-squibs} as a starting point, and use it to expand, review, and refine a paraphrase typology in the next section. By doing so, we aim to provide a more rigorous and scientific approach to understanding paraphrases and their role in natural language processing.

\subsection{Paraphrase Types Overview}
We propose the REfined PARAPHRASE typology definitions (\textsc{ReParaphrased}) as an extension of the Extended Paraphrase Typology (EPT) proposed by \cite{kovatchev_etpc_2018}. \textsc{ReParaphrased} combines the categorization of EPT with non-overlapping paraphrase types proposed by \cite{bhagat-hovy-2013-squibs}, and introduces new paraphrase operations, such as relational substitutions and verbatim paraphrasing. Additionally, the "modal verb changes" in EPT have been expanded to include functional word substitutions, a broader class of paraphrase operations. By providing a more comprehensive and accurate representation of paraphrase types, \textsc{ReParaphrased} enables a deeper understanding of the effect of type distributions on downstream tasks such as paraphrase identification and generation. \textsc{ReParaphrased} contains a total of 24 paraphrase types, as shown in Table \ref{tab:1}.

\begin{table}[!t]
\caption{All the types of paraphrases in \textsc{ReParaphrased}}
\centering
\begin{tabular}{ |c||c|  }
 \hline
 \multicolumn{2}{|c|}{Paraphrase Types} \\
\hline
\multirow{6}{*}{\rotatebox[origin=c]{90}{\parbox[c]{1.5cm}{\centering Morphology Based}}}
&\\
&  Inflectional Changes\\
&  Derivational Changes\\
&  Functional Word substitution\\
&\\
&\\
\hline
\multirow{8}{*}{\rotatebox[origin=c]{90}{\parbox[c]{1cm}{\centering Lexicon Based}}}   
&\\
& Same Polarity Substitution\\
& Opposite Polarity Substitution\\
& Converse Substitution\\
& Spelling Changes\\
& Synthetic / Analytic Substitution\\
& Relational substitution \\
&\\
\hline
\multirow{7}{*}{\rotatebox[origin=c]{90}{\parbox[c]{1cm}{\centering Syntax Based}}} 
&\\
& Negation Switching\\
& Diathesis Alternation\\
& Subordination And Nesting Changes\\
& Coordination Changes\\
& Ellipsis\\
&\\
\hline
\multirow{6}{*}{\rotatebox[origin=c]{90}{\parbox[c]{1cm}{\centering Discourse Based}}} 
&\\
& Syntax / Discourse Changes\\
& Direct/Indirect Substitution\\
& Sentence Modality Changes\\
& Punctuation Changes\\
&\\
\hline
\multirow{8}{*}{\rotatebox[origin=c]{90}{\parbox[c]{1cm}{\centering Other Changes}}} 
&\\
& Change Of Order\\
& Change of Format\\
& Addition / Deletion Changes\\
& Entailment\\
& Verbatim\\
& Identity\\
&\\
\hline
\label{tab:1}
\end{tabular}
\end{table}

\subsection{Paraphrase Types}
\paragraph{Same Polarity Substitutions}
Same polarity substitutions, also known as synonym substitutions, involve the replacement of a word or phrase of the sentence with one of its synonyms. Same polarity substitutions have three subtypes: 1. habitual (replacing a verb with its synonym), 2. contextual (replacing a phrase with an equivalent phrase based on context), and 3. named entities (replacing a noun with its equivalent alternative).
\begin{enumerate}
    \item  I \textbf{dislike} doing extra work. $\Longleftrightarrow$ I \textbf{hate} doing extra work.
    \item Their bank account balance \textbf{reached the maximum insured amount} $\Longleftrightarrow$ Their bank account balance was \textbf{at least 250 thousand dollars} 
    \item \textbf{Mr. Smith} just bought a new computer $\Longleftrightarrow$ \textbf{Bob} just bought a new computer
\end{enumerate}
\paragraph{Opposite Polarity Substitutions}
Opposite Polarity Substitutions are also known as antonym substitutions, where words and phrases in a sentence are replaced by their negated antonyms. There are two types of substitutions in this category: 1. habitual (main verb or adverbs are replaced by one of its antonyms) and 2. contextual (phrases are replaced by their alternative with opposite meanings)  
\begin{enumerate}
    \item The program \textbf{runs fast} $\Longleftrightarrow$ The program \textbf{does not run slowly}
    \item A spike in sales performance will \textbf{save the company from bankruptcy} $\Longleftrightarrow$ Only a spike in sales performance \textbf{will halt the company's bankruptcy}.  
\end{enumerate}
\paragraph{Converse Substitution}
Converse substitution involves the relational substitution of a word in a sentence by its relational pair with an opposite viewpoint. 
\begin{description}
    \item I \textbf{bought} a plane ticket online $\Longleftrightarrow$ A plane ticket was \textbf{sold} to me online
\end{description}
\paragraph{Inflectional Changes}
Inflectional changes involve the inflection of nouns (usually inflected for numbers) and the inflection of verbs (inflected for tense).
\begin{description}
    \item Increase in \textbf{salaries} is often a great indicator of performance $\Longleftrightarrow$ Increase in \textbf{salary} is often a great indicator of performance. 
\end{description}
\paragraph{Sentence Modality Changes}
Sentence modality changes involve the overall change in the expression of perspectives regarding certainly toward the subject of the sentence.
\begin{description}
    \item Does working in that technology company pay well? Does it provide great 401k plans for its employees? $\Longleftrightarrow$ They will work at the company to get high pay or to obtain great 401k plans. 
\end{description}
\paragraph{Functional Word Substitution}
Functional word substitution involves substituting a functional word in a sentence with another functional word. 
\begin{description}
    \item Is \textbf{this} your own work? $\Longleftrightarrow$ Is \textbf{that} your own work?
\end{description}
\paragraph{Spelling Changes}
Spelling changes involve changes through contractions (combining two words) or verb conjugations. 
\begin{description}
    \item The countless hours spent practicing \textbf{did not} improve our performance $\Longleftrightarrow$ The countless hours spent practicing \textbf{didn’t} improve our performance
\end{description}
\paragraph{Structure/Discourse Changes}
Structure and discourse changes involve changes in the referencing context of the discourse in a sentence.
\begin{description}
    \item How he would stare! $\Longleftrightarrow$ He would surely stare!
\end{description}
\paragraph{Relational Substitutions}
Relational substitutions involve the substitution of a word or phrase by its relational counterparts. The relational counterparts are defined in two ways: 1. agent(action)/action(agent) substitution, and 2. manipulator/device substitution.
\begin{enumerate}
    \item Jacob \textbf{programmed} the game $\Longleftrightarrow$ the game's \textbf{programmer} is Jacob
    \item That \textbf{driver} is speeding on the highway $\Longleftrightarrow$ That \textbf{car} is speeding on the highway
\end{enumerate}
\paragraph{Derivational Changes}
Derivational changes involve the change of a verb to its adjective form in a sentence.
\begin{description}
    \item There are many accounts of that hero’s legacy all \textbf{differing} in perspectives. $\Longleftrightarrow$ There are \textbf{different} versions of that hero’s legacy. 
\end{description}
\paragraph{Direct/Indirect Style Alternation}
Direct and indirect style alternation involves the changing of voices in a sentence such as from active to passive voice and vice versa. 
\begin{description}
    \item “You must finish the project by the end of today,” \textbf{demanded my manager} $\Longleftrightarrow$ \textbf{My manager demanded that} I must finish the project by the end of today
\end{description}
\paragraph{Punctuation Changes}
Punctuation changes involve the change of punctuation used in the sentence. 
\begin{description}
    \item These numbers\textbf{, interestingly,} seem to appear in the world around us. $\Longleftrightarrow$ These numbers \textbf{interestingly} seem to appear in the world around us.
\end{description}
\paragraph{Coordination Changes}
Coordination changes involve the connection of two related sentences through the use of conjunctions.
\begin{description}
    \item The most popular sport in the world is Basketball. \textbf{In addition,} it is also the sport that pays its athlete the most. $\Longleftrightarrow$ The most popular sport in the world is Basketball \textbf{, and} it is the sport that pays its athlete the most. 
\end{description}
\paragraph{Ellipsis}
Ellipsis involves the omission of clauses that are understood from the context of the remaining sentence. 
\begin{description}
    \item Alice started the homework a few weeks prior to the deadline but was unable to finish it before the deadline. $\Longleftrightarrow$ Alice started the homework a few weeks prior to the deadline, but \textbf{she} was unable to finish it before the deadline.
\end{description}
\paragraph{Negation Switching}
Negation switching involves changing the negation in a sentence with an equivalent alternative. 
\begin{description}
    \item We need \textbf{not} any complicated equations $\Longleftrightarrow$ We \textbf{do not} need any complicated equations
\end{description}
\paragraph{Diathesis Alternation}
Diathesis alternation, also known as verb alternation, involves the alternation of the arguments of a sentence.
\begin{description}
    \item Alice presented \textbf{the gift to Bob}. $\Longleftrightarrow$ Alice presented \textbf{Bob with the gift}.
\end{description}
\paragraph{Subordination and Nesting Changes}
Subordination and nesting changes involve substituting an element within the sentence by an overarching class that the element belongs to or by a member of that element. 
\begin{description}
    \item All \textbf{spoken languages} are natural languages $\Longleftrightarrow$ The \textbf{English language} is a natural language
\end{description}
\paragraph{Synthetic/Analytic Substitution}
Synthetic and analytic substitution involves elaborating the syntactic attributes of a word or phrase.
\begin{description}
    \item Comments $\Longleftrightarrow$ \textbf{A variety of} comments
\end{description}
\paragraph{Change Of Order}
Change Of Order involves changing the order of a word or phrase in a sentence. 
\begin{description}
    \item \textbf{Initially,} we begin with the scientific method $\Longleftrightarrow$ we begin with the scientific method \textbf{initially}
\end{description}
\paragraph{Change of Format}
Change of format involves changing numerical numbers and symbols to their written counterparts and vice versa. 
\begin{description}
    \item  \textbf{two} hours $\Longleftrightarrow$ \textbf{2} hours
\end{description}
\paragraph{Addition/Deletion Changes}
The addition and deletion changes involve the addition of new details or the deletion of existing details from a sentence.\\ 
\begin{description}
    \item \textbf{Yesterday}, we were able to \textbf{complete our assignment and} submit it on time  $\Longleftrightarrow$ \textbf{Yesterday evening at 12:30 PM}, we were able to submit our assignment on time
\end{description}
\paragraph{Entailment}
Entailment involves substituting a phrase within a sentence with another phase in which the original phrase entails. 
\begin{description}
    \item A highly-regarded company \textbf{bought} its competitors  $\Longleftrightarrow$  A highly-regarded company \textbf{intends to buy} its competitors 
\end{description}
\paragraph{Verbatim}
Verbatim is a type of plagiarism where a sentence is copied without changing any aspect of the sentence.
\begin{description}
    \item \textbf{The algorithm performed well on this dataset }$\Longleftrightarrow$ \textbf{The algorithm performed well on this dataset}
\end{description}
\paragraph{Identity}
Identity closely resembles verbatim, and the main difference between the two is that identity usually copies a subsection of a sentence or phrase. 
\begin{description}
    \item The manager told us that \textbf{we have two more days to complete the project} $\Longleftrightarrow$ Our boss texted in our work channel that \textbf{we have two more days to complete the project}
\end{description}

\section{Paraphrase identification Review}

Paraphrase identification (PI) determines the semantic similarity of two phrases or texts based on quantitative measurements \cite{fujita_expanding_2018,al-smadi_paraphrase_2017}. The methods of PI have been extensively integrated into tasks such as paraphrase recognition, classification, and detection \cite{fujita-sato-2008-probabilistic}. In this section, we classify paraphrase recognition methods based on traditional approaches and deep neural networks as modern approaches. This section briefly reviews traditional techniques and approaches for PI, including knowledge-based and corpus-based methods. Traditional approaches focused on two domains of linguistics: morphology and syntax, which were widely adopted and taken as benchmarks by modern approaches. Then we review modern techniques that leverage deep learning, following the challenges of paraphrase identification with traditional methods. We also provide the main architectures of all methods in Fig. \ref{fig:DLA} for future research.

\subsection{Traditional approaches and techniques for paraphrase identification}

\begin{figure*}[!t]
\centering
\includegraphics[width=5in]{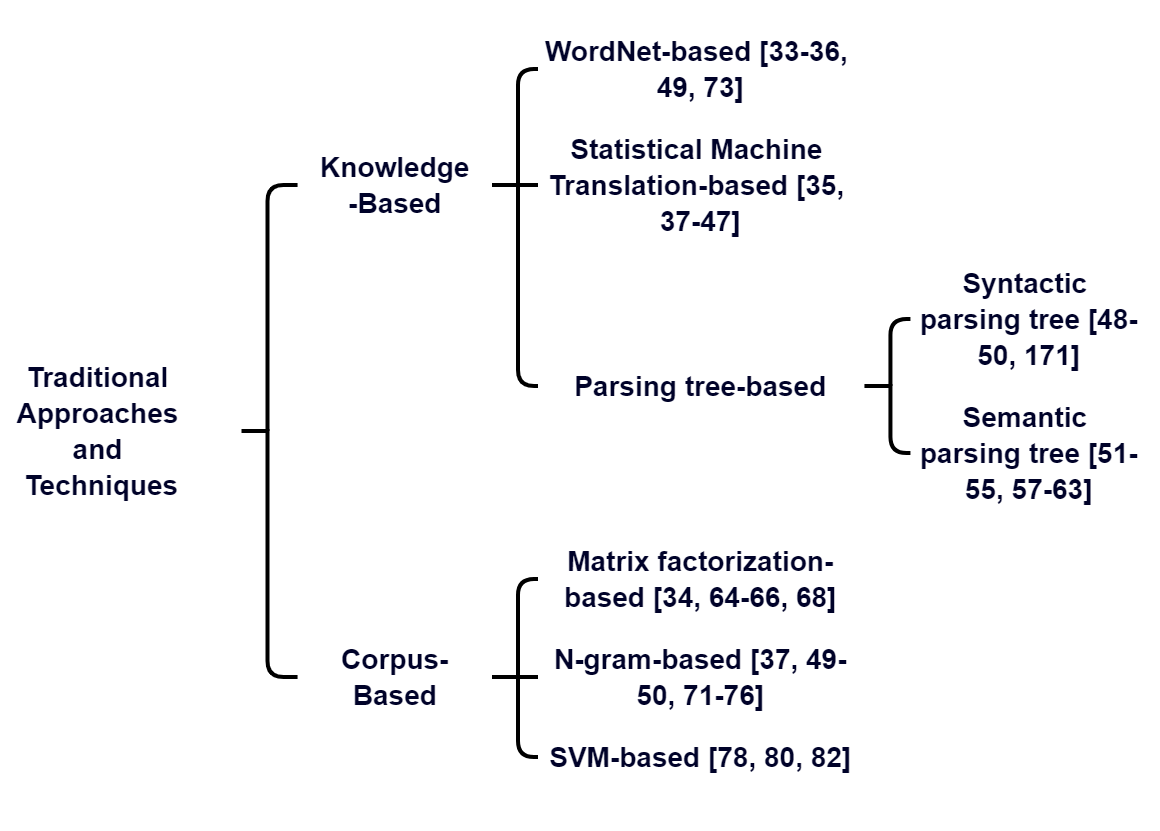}
\caption{Traditional Approaches and Techniques on paraphrase identification are mainly classified as knowledge-based and corpus-based. Each of them has various branches. We review the three most used ones for each category.}
\label{fig:traapptec}
\end{figure*}

Before the advent of deep learning, rule-based and non-neural-network-based methods were applied to PI tasks. These traditional methods are mainly divided into knowledge-based and corpus-based methods. Knowledge-based methods tend to use common sense knowledge from lexical database sources such as dictionaries, WordNet \cite{miller_wordnet_1995}, and dependency trees. Corpus-based measures capture the semantic similarity of words and texts calculated from data rather than compiled knowledge rules. Most corpus-based methods are probabilistic. Researchers often combine techniques from both methods. Figure \ref{fig:traapptec} shows the tree structure of traditional and modern approaches and techniques for PI.

\subsubsection{Knowledge based approaches}

One of the traditional approaches to identifying paraphrases involves using WordNet, a lexical database of semantic relations \cite{miller_wordnet_1995}. In this approach, synonyms and hypernyms from WordNet are used to filter possible paraphrases and reduce the scope of the identification task. For example, in \cite{mihalcea_corpus-based_2006}, researchers used six WordNet-based metrics to explore a text-to-text approach to identifying paraphrases. They transformed concept-to-concept similarity into word-to-word similarity by selecting words in WordNet relationships. In \cite{finch_using_2005}, a modified edit distance algorithm was run on WordNet relationships between words to handle synonyms, resulting in a 0.6\% improvement on a standard semantic equivalence task.

However, the effectiveness of WordNet relationships in identifying paraphrases is limited by their small lexical coverage and lack of vocabulary diversity. In an effort to expand the use of WordNet relationships, \cite{cocos_mapping_2017} mapped the Paraphrase Database (PPDB) to WordNet to predict WordNet synset membership of pairs that did not exist in WordNet. This approach resulted in an accuracy of 89\% when expanding from approximately 155,000 words in WordNet to over 74 million words. Despite this progress, there is still room for further improvement in the use of WordNet-based approaches for identifying paraphrases.

\paragraph{Statistical Machine Translation based approaches}
The use of Statistical Machine Translation (SMT) evaluation has been shown to be effective in identifying semantic paraphrases at the sentence level. In \cite{finch_using_2005}, machine translation evaluation methods such as BLEU \cite{papineni-etal-2002-bleu}, NIST \cite{10.5555/1289189.1289273}, WER \cite{WER-1992}, and PER \cite{Tillmann97accelerateddp} were utilized to build a paraphrase classifier. The authors also employed an SVM to classify paraphrased and non-paraphrased sentences based on the feature vectors produced by the machine translation evaluation systems. This research has inspired other researchers to map MT evaluation approaches to paraphrase identification methods. For example, in \cite{wan_using_2006}, the use of dependency-based features such as Part-of-Speech (POS) resulted in a 4.4\% accuracy improvement and a 6\% F1 improvement.  In \cite{madnani_re-examining_2012}, five novel MT metrics (TER-Plus \cite{10.2307/40783463}, METEROR \cite{denkowski-lavie-2010-extending}, SEPIA \cite{habash2008sepia}, BADGER \cite{Parker_badger:a}, and MAXSIM \cite{chan-ng-2008-maxsim}) were tested on the MSRPC dataset. The results showed that using MT evaluation metrics alone had good performance on the MSRPC dataset. However, the use of MT evaluation metrics alone may not always be effective in identifying paraphrases, especially in complex datasets such as the Plagiarism Detection Corpus (PAN) \cite{potthast2009pan}.

\paragraph{Parsing tree-based approaches}

One approach to identifying semantic paraphrases is the use of parsing trees, which compare texts by their underlying tree structures that represent them. Syntactic parsing trees, such as those built using the Penn Treebank dataset \cite{taylor_penn_2003}, are constructed using POS tagging, syntactic bracketing, and disfluency annotation schemes. The Penn Treebank has significantly influenced NLP research, and has been used to train syntactic classifiers for paraphrase identification. For example, in \cite{das_paraphrase_2009}, the authors showed that the Penn Treebank can be used to build improved English dependency parsing models, leading to an increase in the quality of paraphrase detection. In \cite{sidorov_syntactic_2013}, the authors combined $n$-gram features with syntactic features from the dependency tree to improve paraphrase detection.

Another type of parsing tree is based on semantic features, known as semantic parsers. These parsers are designed to construct the meaning behind a given sentence. Most previous research in this area \cite{wong-mooney-2006-learning, zettlemoyer-collins-2009-learning, sun2014empirical} has relied on large amounts of human annotation to build semantic parsers. However, this can be time-consuming and expensive. To address this, \cite{bao_knowledge-based_2014} proposed a weakly supervised semantic parsing method that does not require fine-grained annotations. \cite{dong_statistical_2015} presented a translation-based weakly supervised semantic parsing method that translates questions into answers.

One approach that has gained traction in recent years is the use of Abstract Meaning Representation (AMR) to assign precise semantic representations to sentences with the same meaning \cite{bouamor_multitechnique_2013}. Approaches based on AMR have evolved from depending on monolingual datasets to using multilingual datasets, and have achieved state-of-the-art performance on this task \cite{flanigan_discriminative_2014, peng_synchronous_2015, zhou_amr_2016, konstas_neural_2017, issa_abstract_2018, blloshmi_xl-amr_2020, ribeiro_structural_2021}. The use of AMR allows for the effective identification of paraphrases across languages and has proven to be a powerful tool in the field of semantic paraphrase identification.

\subsubsection{Corpus-based approaches}
\paragraph{Matrix factorization-based approaches}
Matrix factorization is a common technique for reducing the dimensionality of matrices in semantic paraphrase identification tasks. Singular value decomposition (SVD) is the most widely used matrix factorization method in this context. Latent semantic analysis (LSA), first proposed by \cite{doi:10.1080/01638539809545028}, is a corpus-based measure that can be used for semantic similarity by using SVD to reduce the dimensionality of the term-document matrix representing the corpus. This decomposition can be seen as a by-product of the term co-occurrence matrix in a corpus. For example, in \cite{mihalcea_corpus-based_2006}, the authors apply LSA word similarity measures on a pseudo-document text representation, combining it with a TF-IDF weighting scheme. In \cite{guo-diab-2012-modeling}, the authors show further improvement when unseen words are given additional weight. These methods generally treat sentences as pseudo-documents in an LSA framework and identify paraphrases using similarity in the latent space.

In \cite{ji_discriminative_2013}, the authors propose an improved discriminative term-weighting metric (TF-KLD) inspired by LSA approaches, combined with SVM. Similar to another data mining approach, linear discriminant analysis (LDA), the basic intuition behind this method is to use SVD to perform factorization on the co-occurrence matrix, with the addition of a non-negativity constraint \cite{10.5555/3008751.3008829} in the latent representation based on non-orthogonal basis. Unlike previous related methods, this approach increases the weights of discriminative distributional features while decreases the weights of features that are not. To re-weight the features in the matrix, the authors apply KL-divergence before the matrix factorization. This contribution has inspired other researchers to explore other features and supervised classification approaches in matrix factorization-based models.

One limitation of TF-KLD models is the inability to define weights for words that do not occur in the training corpus. To address this, \cite{yin_discriminative_2016} propose a new scheme called TF-KLD-KNN, which is based on TK-KLD and aims to solve the problem of unseen words by computing the weight of the unknown unit as the average of the weights of its k nearest neighbors. This approach allows for the effective identification of paraphrases even when the words in the sentences do not occur in the training corpus.

Overall, matrix factorization-based approaches have proven to be effective in identifying semantic paraphrases, particularly when combined with other techniques, such as SVM and KL-divergence. These methods have been applied to a wide range of datasets and have shown promising results in identifying paraphrases across languages and domains.

\paragraph{N-gram based approaches}
The concept of $n$-grams was first introduced by \cite{Mar13} in the form of a Markov chain, and was later applied to communication by \cite{shannon1948mathematical}. Researchers have used $n$-grams as a feature-based approach in semantic paraphrase identification tasks. In the early days of research in NLP, \cite{1454428} used $n$-grams in speech recognition systems, and \cite{gavalda-waibel-1998-growing-semantic} applied $n$-grams in their Natural Language Understanding (NLU) system.

More sophisticated $n$-gram modifications have been used in paraphrase identification tasks. Bilingual evaluation understudy (BLEU) \cite{papineni-etal-2002-bleu}, which compares the $n$-grams of a candidate text with the $n$-grams of a reference translation, has become a widely used method for evaluating paraphrases.

In \cite{das_paraphrase_2009}, the authors proposed a logistic regression model incorporating surface features from $n$-grams as a probabilistic lexical overlap model (e.g., the precision and recall of $n$-gram overlaps between two target paraphrases), and combined it with hidden loose word alignment approaches (following \cite{smith-eisner-2006-quasi}, they loosely matched the nodes of the syntactic trees of the two targets). They also followed \cite{wang2007jeopardy} in treating the correspondences as latent variables and using a WordNet-based lexical semantics model to generate the target words.

Recent approaches based on the $n$-gram idea have sought to improve upon this method by incorporating syntactic advantages. For example, \cite{sidorov_syntactic_2014, calvo_dependency_2014, sidorov_syntactic_2013} proposed syntactic $n$-grams, which do not use text directly but rather rely on syntactic dependency analysis trees. However, these methods still face the challenge of scaling well to longer sequence lengths, as \cite{10.5555/1214993} have shown that $n$-grams can be deceived by lexical and syntactic overlaps.

\paragraph{Support Vector Machine-based approaches}
In 2005, Microsoft researchers proposed a novel paraphrase dataset \cite{brockett_support_2005}. They used an annotated dataset of clustered news articles from \cite{dolan_unsupervised_2004} and hand-annotated pairs of sentences judged to be paraphrased. To make the paired comparison more relevant, they used a Support Vector Machine (SVM) to refine the labeling process. This work also showed that SVMs can be effectively used for paraphrase identification, as they are robust to noise in training data and can handle sparse features from paraphrase detection tasks. Moreover, an appropriate choice of an SVM kernel function can prevent overfitting in low-resource datasets.

The fundamental idea behind SVM-based paraphrase identification is to use language entities (e.g., letters, words, sentences) in classification tasks to determine the similarity that is considered paraphrasing. For example, \cite{park_sentential_2014} improved SVMs with a special string-focused kernel \cite{CC01a} to measure the similarity of non-fixed size feature vectors in the language model. Others have adapted the more traditional Radial Basis Function (RBF) to effectively detect paraphrases in large datasets. \cite{eyecioglu_twitter_2015} adapted kernels that fit large datasets with a small number of features. While SVMs are widely used in paraphrase identification, high dimensionality and sparseness in real datasets make the use of SVM in real-world scenarios impractical.

\subsection{Paraphrase Identification with Deep Learning}
Paraphrase identification is a fundamental task in natural language processing (NLP). Distributed representations of words, high dimensional real-valued vectors, were proposed decades ago (\cite{rumelhart_learning_1986, Deerwester1990IndexingBL, Neural-Pro-LM-2003, doi:10.1126/science.1127647}). Traditional distributed representations were based on unsupervised statistical language models, such as $n$-gram overlaps. However, these overlaps can be time-consuming to compute, a fact that led to the development of neural network structures that can predict word embeddings directly (e.g., \cite{bengio_greedy_2006, Kombrink_Stefan_Mikolov_2011}). Neural probabilistic language models were introduced to address the curse of dimensionality by learning a distributed representation of words (\cite{Neural-Pro-LM-2003}). This allowed for the training of large models with millions of parameters within a reasonable time. The distributed representations can be used for paraphrase identification tasks based on their similarity, and researchers are continuing to find more accurate and efficient methods based on this idea.

\begin{figure*}
    \begin{center}
        \includegraphics[width=5in]{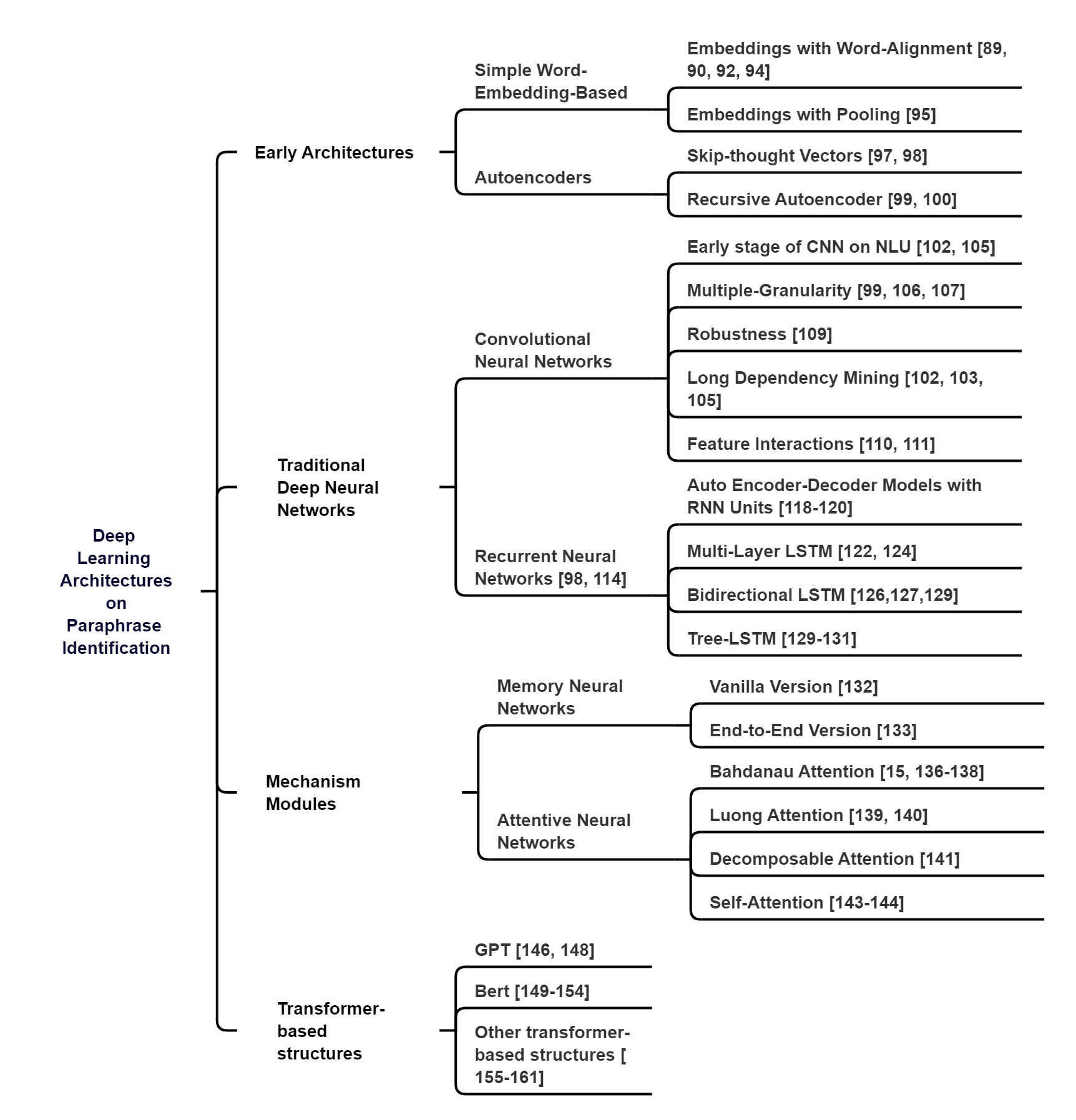}
        \caption{Early Architectures consist of simple embedding approaches and shallow neural networks on PI tasks. Traditional deep neural networks (TDNNs) contain two mainstream neural networks: CNNs and RNNs, and their improvements on PI tasks. Mechanism Modules include substantial independent improvements on PI tasks. Transformer-based structures incorporate modern transformers on PI or downstream tasks.}
        \label{fig:DLA}
    \end{center}
\end{figure*}

Techniques and approaches initially designed for downstream tasks, such as question answering (QA), recognize text entailment (RTE), natural language inference (NLI), and semantic text similarity (STS), can also be applied to paraphrase identification tasks. In this section, we introduce mainstream approaches that have been applied to paraphrase identification tasks. Deep learning approaches to paraphrase identification often have many overlapping intersections. We also provided Table III which describes the challenges of each model. Figure \ref{fig:DLA} shows the relationships between the works reviewed.

\begin{table*}
\begin{center}
\caption{Deep Learning Models' Research Challenges on PI}
\begin{tabular}{ p{0.7cm} p{0.5cm} p{1.3cm} p{7.3cm} p{2cm} p{1.5cm}}
 \hline
Model & Year & Structures &  Challenges of PI & Learning Method & Datasets \\
 \hline
\cite{blacoe-lapata-2012-comparison} & 2012 & SWEMs & Compositional Distributional Semantics (Word Level) & Unsupervised & WordSim353 \\
\cite{banea_simcompass_2014} & 2014 & SWEMs & Compositional Distributional Semantics (Word Level) & Unsupervised & Sem 2013\\
\cite{yu2014deep} & 2014 & SWEMs & Compositional Distributional Semantics (Word Level) & Unsupervised & TREC \\
\cite{he_pairwise_2016} & 2016 & Hybrid & Compositional Distributional Semantics (Word Level) & Supervised & SICK et al.\\
\cite{rocktaschel2016reasoning} & 2016 & RNNs & Compositional Distributional Semantics (Word Level) & Supervised & SNLI \\
\cite{shen_baseline_2018} & 2018 & SWEMs & Compositional Distributional Semantics (Word Level) & Supervised & Yahoo et al.\\
\cite{tan2018multiway} & 2018 & RNNs & Compositional Distributional Semantics (Word Level) & Supervised & SNLI et al.\\
\cite{socher_dynamic_2011} & 2011 & RAEs & Compositional Distributional Semantics (Phrase Level) & Hybrid & MSRP \\
\cite{huang2011paraphrase} & 2011 & RAEs & Compositional Distributional Semantics (Phrase Level) & Unsupervised & MSRP et al.\\
\cite{DBLP:journals/corr/ChoMGBSB14} & 2014 & RNNs & Compositional Distributional Semantics (Phrase Level) & Supervised & WMT’14 \\
\cite{Synactic-tree-LSTM-2015} & 2015 & RNNs &  Compositional Distributional Semantics (Phrase Level) & Supervised & SemEval2014\\
\cite{peng2022towards} & 2022 & Transformer & Compositional Distributional Semantics (Phrase Level) & Supervised & PAWS et al.\\
\cite{DBLP:journals/corr/KirosZSZTUF15} & 2015 & Hybrid & Compositional Distributional Semantics (Sentence Level) & Unsupervised & SICK \\
\cite{Hu-et-al-2014} & 2014 & CNNs & Compositional Distributional Semantics (Sentence Level) & Supervised & MSRP \\
\cite{ABCNN-yin-2015} & 2015 & CNNs & Compositional Distributional Semantics (Sentence Level) & Supervised & MSRP et al.\\
\cite{Multipossen2016} & 2016 & RNNs & Compositional Distributional Semantics (Sentence Level) & Supervised & QA\\
\cite{tay2018co} & 2018 & RNNs & Compositional Distributional Semantics (Multiple Granularity) & Supervised & SNLI et al.\\
\cite{Collobert_and_Weston_2008} & 2008 & CNNs & Compositional Distributional Semantics (Multiple Granularity) & Semi-Supervised & PropBank \\
\cite{yin_convolutional_2015} & 2015 & CNNs & Compositional Distributional Semantics (Multiple Granularity) & Unsupervised & MSRP\\
\cite{DiSAN-Tao-2017} & 2017 & Self-attention & Compositional Distributional Semantics (Multiple Granularity) & Supervised & SICK et al.\\
\cite{he-etal-2015-multi} & 2015 & CNNs & Compositional Distributional Semantics (Multiple Granularity) & Hybird & MSRP et al.\\
\cite{wang2017bilateral} & 2017 & RNNs & Compositional Distributional Semantics (Multiple Granularity) & Supervised & Quora\\
\cite{gong2017natural} & 2017 & Hybrid & Compositional Distributional Semantics (Multiple Granularity) & Supervised & MultiNLI \\
\cite{ARASE2021101164} & 2021 & Transformer & Compositional Distributional Semantics (Multiple Granularity) & Supervised & PAWS et al.\\
\cite{agarwal_deep_2018} & 2018 & CNNs & Robust Training & Supervised & MSRP et al.\\
\cite{tomar-etal-2017-neural} & 2017 & Decomposable Attention & Robust Training & Supervised & Quora\\
\\
\hline
\end{tabular}
\label{tab:2}
\end{center}
\end{table*}

\subsubsection{Compositional Distributional Semantics (Word Level)}
The main challenge of paraphrase identification is the semantic similarity between two candidate sentences. Compositional distributional semantics provides a computationally efficient way to get the semantic representation of a whole sentence by combining the meanings of individual words in it. Word-level features, also an important part of local information, are crucial for fine-grained meaning extraction. While traditional methods discussed above have introduced related concepts and did some experiments on extracting semantics, the mainstream solution of semantic similarity identification was inspired by the introduction of word embedding. Word embeddings trained by neural networks have been shown to capture syntactic and semantic regularities in language \cite{mikolov_linguistic_2013}. A specific model for computing word embeddings using neural networks was proposed in \cite{mikolov2013efficient}, introducing Word2Vec. Word2Vec embeddings build upon not only the n-gram features of sentences but also higher-level correlations between words and their contexts. Word embeddings such as Word2Vec and others (e.g., GloVe \cite{pennington2014glove}) provided a new research direction for the PI task, achieving high performance because they are computationally efficient and reasonable on meaning composition.

There are two mainstreams of word-level challenge solutions: Simple Word-Embedding-Based Models (SWEMs) and word-alignment-related techniques. Simple Word-Embedding-Based Models (SWEMs) is a solution that relies on finding and optimizing compositional combinations by using word embeddings. Moving beyond simple word embeddings can be a challenge. One naive method is to average the word embeddings that make up a sentence, which is an order-insensitive method. Early attempts to apply this to paraphrase identification focused mainly on word embeddings and word-alignment methods. For example, \cite{blacoe-lapata-2012-comparison} forms sentence representations by summing up word embeddings. Similarly, \cite{banea_simcompass_2014} applied these two ideas in SemEval-2014 to compare the semantic similarity of sentences and achieved the highest overall performance in the competition. Building on the work of \cite{mihalcea_corpus-based_2006}, which used improved word alignment and similarity measures, \cite{yu2014deep} proposed enhancements to word embeddings and alignments for the PI task. However, this approach was only slightly improved due to its inherent order insensitivity. More complex corpora and semantic features are needed to consider word order. Despite this, word embeddings provided a strong foundation for later research on paraphrase identification using neural networks. One improved idea based on SWEMs was proposed by \cite{shen_baseline_2018}. In their work, they improved SWEMs with associated pooling mechanisms. They combined max pooling and hierarchical pooling (i.e., the hierarchy given by the tree structure of semantic parsing) to solve the order-insensitive challenge and incorporated some handcrafted modifications. Although they did not achieve SOTA performance on semantic test similarity tasks, their simple structure is still competitive compared with other neural network approaches. To resolve the word order issue of SWEMs, \cite{shen_baseline_2018} introduced a hierarchical pooling layer to keep track of the local spatial information of a text sequence. Instead of learning the local spatial information via count features, they learn fixed-length representations for the n-grams that appear in the corpus.

Another solution to this challenge is using word alignment or word-to-word attention techniques to enhance the models' performance in getting fine-grained word-level semantic correspondences. For example, \cite{he_pairwise_2016} added a pairwise word interaction layer to capture the semantic correspondences. The approach focuses on the interaction of words and selects the important word interaction for similarity measurement. Their model alleviates the issue of local information missing in coarse-grained sentence semantic models. At the same time, the attention mechanism has been proven efficient in capturing word-level features because it fixes the limitation of Long Short-Term Memory (LSTM) cell state---a very common recurrent neural network design that suffers from not being able to capture long-term dependencies. \cite{rocktaschel2016reasoning} applies word-by-word attention to a recognizing textual entailment (RTE) issue. Like the paraphrase identification task, they do not use attention to generate words but to obtain a sentence-pair encoding from fine-grained reasoning via soft alignment of words. Their ablation test showed that word-by-word attention yielded another 1.2 percent improvement and empirically shows it's more reasonable in capturing the keywords in a sentence. An advanced approach to solving the word-alignment challenge proposed by \cite{tan2018multiway} was inspired by multiway attention. They concatenated two widely-used attention to model the interactions between sentences and proposed two other functions to calculate the word relation by the element-wise dot product and difference of two vectors. The matching-aggregation framework of their work combined matching information by two Bi-LSTMs and passed them through an attention-pooling layer to a multilayer perceptron for the final decision. These word alignment and attention-based approaches have significantly improved the ability of models to capture fine-grained semantic correspondences, addressing a key challenge in paraphrase identification at the word level.

\subsubsection{Compositional Distributional Semantics (Phrase Level)}
Phrase-level compositional distributional semantics constitutes another facet of local information, which has predominantly been explored within the context of recursive autoencoder-decoder architectural models that operate based on syntactic parse trees inherent in sentences. In Paraphrase Identification (PI) models, phrase-level attributes are typically construed as either aggregation of words or sub-structures intrinsic to sentences. This avenue is particularly pursued when coarse-grained sentence-level attributes fall short of adeptly encapsulating semantic nuances. Illustratively, recursive autoencoder models designed for paraphrase identification learn phrase attributes for every node present in a phrase tree. \cite{socher_dynamic_2011} put forth an extended Recursive Autoencoder (RAE) model, tailored for assimilating feature vectors pertaining to phrases in syntactic trees. This innovation yielded impressive accuracy across both syntactic and semantic constructs. Their introduced dynamic pooling layer, responsible for computing fixed-size representations from variable inputs, alongside an iterative RAE algorithm facilitating the reconstruction of all nodes in the decoder, culminated in a comprehensive phrase-level representation. In parallel, \cite{huang2011paraphrase} adopted the RAE model to acquire representations for the entire input sentence and its constituent sub-phrases, derived from parse trees. Their subsequent utilization of a Support Vector Machine (SVM) to classify aggregated features from the output exhibited proficient performance in discerning concise paraphrases. Concurrently, an alternative avenue was explored by \cite{DBLP:journals/corr/ChoMGBSB14}, who postulated that the extraction of phrase-level features can be enhanced through scored phrase pairs. They proposed an RNN encoder-decoder model and seamlessly integrated it into a standard phrase-based Statistical Machine Translation (SMT) system. The integration involved scoring each phrase pair within the framework. Empirical assessments showcased a marked improvement in the extraction of meaning with the integration of scored phrase pairs. It's important to underscore, however, that these models operating at the phrase level necessitated extensive manual engineering efforts to achieve commendable performance \cite{socher_dynamic_2011}.

In recent times, strides have been taken to enhance phrase alignment techniques within transformer-based models. To exemplify, \cite{peng2022towards} introduced a phrase alignment algorithm encompassing a BERT-based Semantic Role Labeling (SRL) tagger as a precursor to alignment. Subsequently, they adapted the Jonker-Volgenant algorithm to effectuate optimal phrase alignment. This approach reduced the requirement for labor-intensive manual crafting and introduced a novel approach to tackling the challenges inherent in phrase-level tasks.

\subsubsection{Compositional Distributional Semantics (Sentence Level)}
Sentence-level challenges in compositional distributional semantics focus on linking "linguistic entities" at the sentence level. These include both the internal structure of sentences and how they relate to each other. Most traditional deep learning models for Paraphrase Identification (PI) look at sentences separately, missing out on how they connect. To fix this, \cite{DBLP:journals/corr/KirosZSZTUF15} created the skip-thought vector model. This model combines the skip-gram approach with Gated Recurrent Units (GRUs) in an encoder-decoder setup. It takes the word embedding idea from the skip-gram model \cite{mikolov2013efficient} and applies it to whole sentences, capturing their context. Unlike phrase and word-level methods, skip-thought doesn't need extra handmade features. It performs well against recursive networks that use dynamic pooling for PI tasks. However, it still struggles with complex texts because it uses broad sentence features that do not capture all the semantic details.

\begin{figure*}[!t]
\centering
\includegraphics[width=5in]{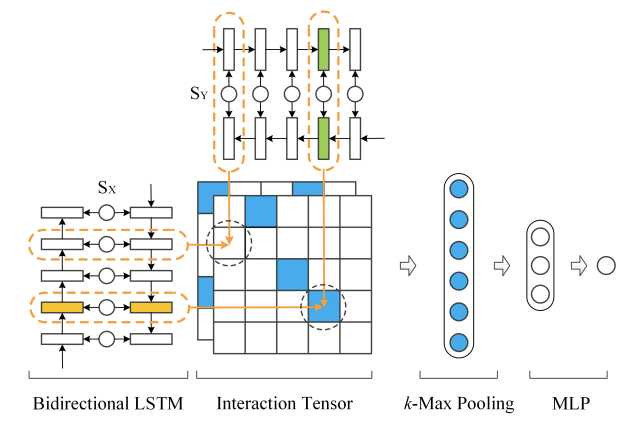}
\caption{The Tensor layer and k-Max pooling mechanism in \cite{Multipossen2016}. This figure cites from \cite{Multipossen2016}. }
\label{fig:multi01}
\end{figure*}

In their effort to extract sentence-level features, \cite{Hu-et-al-2014} developed a multi-faceted approach that examines both the internal structure of sentences and how they interact. Moving away from recursive models like Recurrent Neural Networks (RNNs) and Recursive Autoencoders (RAEs), their Convolutional Neural Network (CNN) model avoids single-path compositions from syntactic parsing. Instead, it uses max-pooling to create a wide feature map with multiple selections. Their ARC-I architecture employs a multi-layer perceptron (MLP) to analyze sentence interactions. However, ARC-I has a major flaw: it cannot capture sentence interactions during the forward pass, only comparing sentences after their individual representations are fully formed. This results in lost details about how the sentences interact. To fix this problem, they created the ARC-II architecture, which applies 2D convolution to sentence interactions before the MLP layer. The key innovation here is expanding 1D convolution, typically used for examining sentence structure, into 2D convolution that can analyze relationships between sentences. Building on this work, \cite{Multipossen2016} introduced a tensor layer for interaction and combined it with K-Max pooling. This approach picks out the most important interactions from each matrix, as shown in Fig. \ref{fig:multi01}.

A different approach to sentence-level composition was proposed by \cite{ABCNN-yin-2015}. They introduced attention mechanisms to incorporate mutual influence into Convolutional Neural Networks (CNNs). Their innovative method uses attention feature matrices to adjust convolution and pooling. The first attention feature matrix comes from matching the representation matrices of the two sentences. It acts as an extra feature map added to the convolutional layer along with the representation matrices. This attention feature matrix guides the convolution process towards "counterpart-biased" features, which are similar to mutual information. After that, they create a second attention matrix from the convolution output, which improves the convolution features. They added this second attention to help with better pooling when filtering larger-scale features. Despite these advancements, earlier sentence-level composition methods still struggle to align specific parts of two potential paraphrases.

\subsubsection{Compositional Distributional Semantics (Multiple Granularities)}
The complex challenges of varying granularity require Paraphrase Identification (PI) models to handle features at multiple levels, rather than focusing on just one level, as we discussed earlier. The Recursive Autoencoder-Decoder (RAE) model is a good example of this approach. It deals with this challenge by calculating representations at all levels of a parse tree \cite{socher_dynamic_2011, huang2011paraphrase}. However, RAE models rely heavily on parsing trees, which are not always available for PI tasks. Another way to tackle these multi-level granularity problems is to use convolutional methods to extract multi-level features through stacked Convolutional Neural Networks (CNNs) or pooling techniques \cite{Collobert_and_Weston_2008, yin_convolutional_2015, kalchbrenner-etal-2014-convolutional}. For example, \cite{Collobert_and_Weston_2008} created a single CNN architecture for multitask learning. To address the multi-granularity issue, they used stacked Time-Delay Neural Networks (TDNNs) with convolution, which extract local information at lower levels and global insights at higher levels. Building on this, \cite{kalchbrenner-etal-2014-convolutional} suggested $k$-max pooling to handle multiple granularities. This method captures the $k$ most important features, allowing the model to abstract higher-order and long-range features. Additionally, the Max-TDNN model, with its built-in subgraphs, effectively captures broad semantic connections between words that have few syntactic similarities. As a result, this structure works well with difficult-to-parse texts like Tweets or short messages.

\begin{figure*}[!t]
\centering
\includegraphics[width=5in]{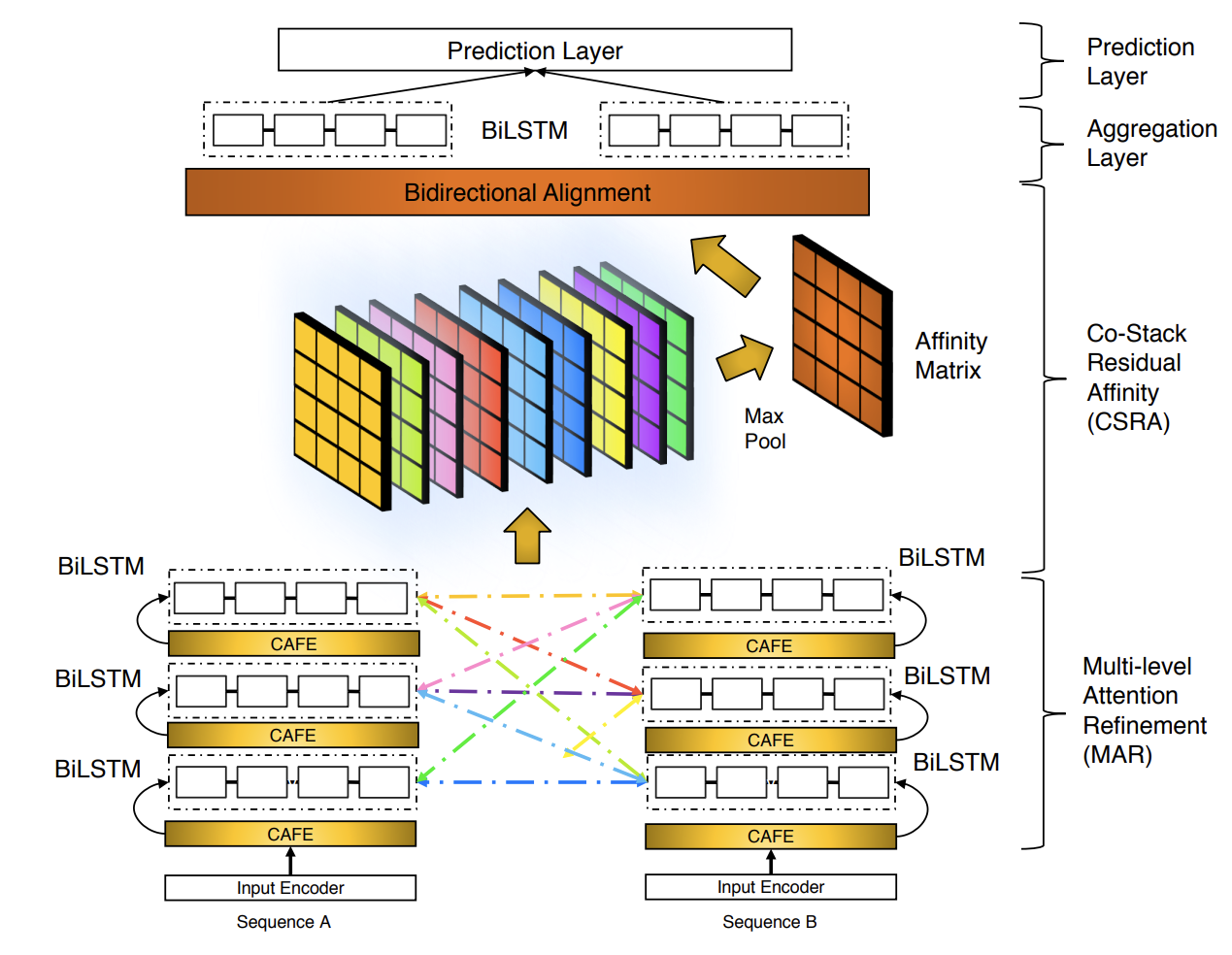}
\caption{Illustration of the proposed Co-Stack Residual Affinity Network (CSRAN) architecture\cite{tay2018co}. Each color-coded matrix represents the interactions between two layers of sequence A and sequence B. This figure cite from \cite{tay2018co}. }
\label{fig:multi02}
\end{figure*}

Yin et al. \cite{yin_convolutional_2015} introduced a novel approach with their BI-CNN-MI model. This model analyzes interactions between sentences using two separate Convolutional Neural Networks (CNNs), then condenses the features using logistic regression. They employ dynamic k-max pooling, which uses various matrices including unigram similarity, short n-gram similarity, long n-gram similarity, and sentence similarity. This approach enhances the system's overall performance. Their method allows for the concurrent use of different feature matrices in a single convolution layer, with each matrix extracting unique sentence characteristics simultaneously. In a related development, Tay et al. \cite{tay2018co} presented co-stack residual affinity networks (CSRAN) as a solution to this challenge. Their model incorporates a bidirectional alignment mechanism that determines affinity weights by combining sequence pairs across stacked hierarchies. It also includes a multi-level attention refinement component between stacked recurrent layers. This architecture improves gradient flow and captures features at multiple levels of granularity by utilizing information across all network hierarchies. Figure \ref{fig:multi02} illustrates the structure of this architecture.

Transformer-based models leverage their architecture, attention mechanisms, and training methodologies to comprehend multiple semantic levels and textual relationships. BERT \cite{devlin_bert_2019} introduced masked language modeling, which excels at understanding context from both preceding and succeeding words. This capacity to capture varied contextual layers makes it particularly valuable for tasks involving sentence-level semantics, such as paraphrase identification. Expanding on BERT, \cite{ARASE2021101164} proposed a transfer fine-tuning approach using phrasal paraphrases. This method enhances BERT's ability to assess semantic equivalence between sentences without increasing model size. ALBERT \cite{lan2020albert}, a BERT variant, aims to optimize the balance between model dimensions and training duration. It employs factorized embedding parameterization to facilitate learning of dependencies across words and phrases. T5 \cite{raffel2020exploring} adopts a distinct approach, recasting all Natural Language Processing (NLP) tasks as text-to-text problems, where both inputs and outputs are in textual format. This transformation enables T5 to address diverse granularities by embedding different information tiers in inputs and subsequently decoding relevant insights in outputs. ELECTRA \cite{clark2020electra} introduces a novel training regime centered on predicting substituted tokens within sentences. This innovation empowers the model to capture intricate relationships and semantic nuances across various textual segments.

\subsubsection{Paraphrase Identification Robust Training}
Robust training data plays a crucial role in paraphrase identification, as evidenced by trends in previous methodologies. This aspect typically relates to a model's ability to generalize. These approaches have heavily relied on carefully curated datasets, such as the widely used Microsoft Paraphrase Corpus (MSRP) \cite{dolan_automatically_2005}. However, they often struggle when faced with the complex task of identifying user-generated paraphrases, a challenge rooted in the diverse nature of linguistic variation, including acronyms and specialized emojis. A clear example of this limitation is seen in conventional paraphrase identification methods based on matrix factorization, as shown in the work of \cite{ji_discriminative_2013}. Their approach, while performing well on carefully curated datasets, falters when applied to more complex corpora, failing to accurately identify paraphrases in the SemEval dataset. This stark contrast highlights the gap between controlled experimental settings and the dynamic, often unique language use in real-world contexts. Researchers have integrated attention mechanisms into paraphrase identification. A notable example is the work by \cite{dey-etal-2016-paraphrase}, where they employed attention mechanisms to mitigate noise in paraphrase detection, particularly in datasets from platforms like Twitter. Despite these efforts, results have been modest, revealing the complex interplay between attention mechanisms and the intricacies of informal language.

\begin{figure*}[!t]
\centering
\includegraphics[width=5in]{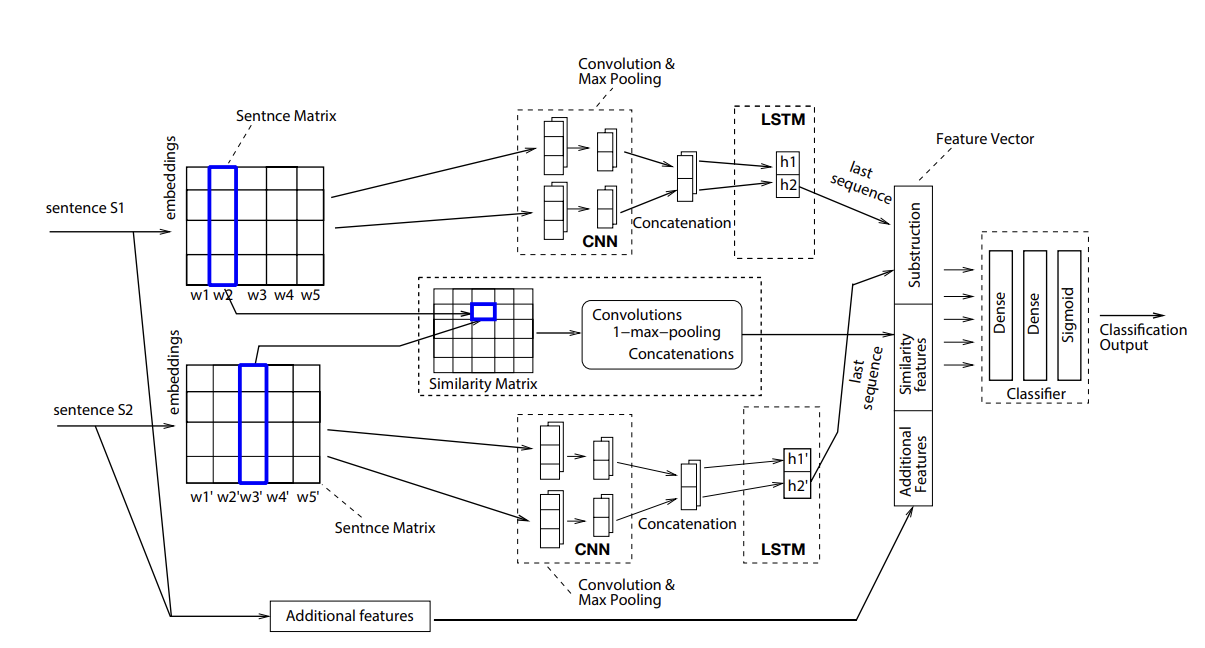}
\caption{The hybrid deep neural architecture for robust paraphrase identification model by \cite{agarwal_deep_2018}. The architecture visually encapsulates the symbiotic fusion of CNNs and LSTMs, exemplifying the innovation's prowess. This figure is cited from \cite{agarwal_deep_2018}.}
\label{fig:robust1}
\end{figure*}

Agarwal et al. \cite{agarwal_deep_2018} introduced a novel hybrid deep neural architecture that advanced the training of robust Paraphrase Identification (PI) models. Their approach combined two key elements: Convolutional Neural Networks (CNNs) for pair-wise word similarity matching, and Long Short-Term Memory networks (LSTMs) for comprehensive sentence modeling. This design effectively captures both broad sentence-level features and specific word-level details, enabling accurate paraphrase detection in noisy, brief Twitter posts. The model's strength lies in its ability to integrate semantic similarity information at both sentence and word levels, allowing it to grasp semantic subtleties across different scales. When faced with grammatical errors or extremely short texts, the word-level similarity component provides crucial insights. In other situations, the sentence's semantic representation becomes more relevant. Figure \ref{fig:robust1} presents a visual representation of this innovative model architecture.

Tomar et al. \cite{tomar-etal-2017-neural} proposed an alternative approach to robust Paraphrase Identification (PI) using a modified version of the decomposable attention model. This model breaks down the prediction process into three key stages: Attend, Compare, and Aggregate. In essence, the model aligns two elements using a neural attention variant, which leads to comparisons between aligned phrases. These aligned phrases are then individually evaluated through a feedforward network. The model subsequently combines the outputs via summation, employing an additional feedforward network and a linear layer, which ultimately yields the label prediction. A significant innovation introduced by the authors involves noisy pretraining. This method entails an initial training phase for all model parameters. This preliminary training uses a relatively small corpus consisting of examples automatically collected from the relevant domain. It is worth noting that these examples inherently contain a certain level of noise due to their automatic acquisition method.

\begin{figure*}[!t]
\centering
\includegraphics[width=5in]{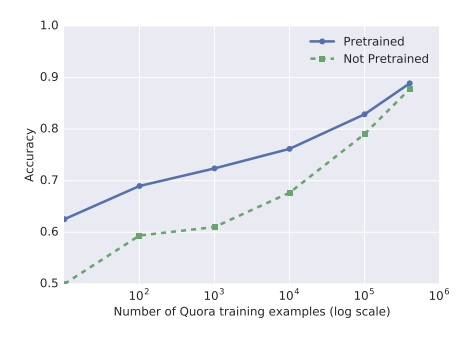}
\caption{Learning curves for the Quora development set with and without pretraining on Paralex by using a noisy pretraining method from \cite{tomar-etal-2017-neural}. This figure is cited from \cite{tomar-etal-2017-neural}.}
\label{fig:robust2}
\end{figure*}

The Transformer model \cite{attention-is-all-2017} has addressed the challenge of robust training in the paraphrase identification task through its unique architecture and mechanisms that enhance its generalization capabilities. It has been widely acknowledged that robust training is crucial for handling linguistic diversity and noise, and the Transformer's self-attention mechanism and positional encodings contribute significantly to this aspect. For instance, consider the sentence: "The conference took place in New York." In this sentence, the position of "New York" is crucial for understanding the location of the conference. The Transformer's positional encodings help the model correctly associate the words with their respective positions, enhancing its robustness in handling diverse input sentences.

\begin{table*}
\begin{center}
\caption{Deep learning methods performance of PI on MSRPC}
\begin{tabular}{ c c c c c c}
 \hline
Model & Year & Encoder &  Embeddings & Accuracy & F1 \\
 \hline
\cite{Hu-et-al-2014}&2014&CNN+MLP+Pooling&Unspervised Embeddings&69.9&80.91\\
\cite{shen_baseline_2018}&2018&SWEM+Pooling&GloVe&71.5&81.3\\
\cite{DBLP:journals/corr/KirosZSZTUF15}&2015&RNN+GNU&Word2Vec&73&82\\\
\cite{blacoe-lapata-2012-comparison}&2012&NLM-BNC&\cite{Collobert_and_Weston_2008}&73.5&82.33\\
\cite{socher_dynamic_2011}&2011&RAE&\cite{Collobert_and_Weston_2008}&76.8&83.6\\
\cite{madnani_re-examining_2012}&2012&Metrics-based&None&77.4&84.1\\
\cite{yin_convolutional_2015}&2015&Bi+CNN+MI&\cite{Collobert_and_Weston_2008}&78.1&84.4\\
\cite{agarwal_deep_2018}&2018&CNN+LSTM+Pooling&Word2Vec&77.7&84.5\\
\cite{ji_discriminative_2013}&2013&Factorization-based&None&80.41&84.59\\
\cite{he-etal-2015-multi}&2015&CNN+Pooling&POS&78.6&84.73\\
\cite{ABCNN-yin-2015}&2015&Bi+CNN+Attention&Word2Vec&78.9&84.8\\
 \hline
\end{tabular}
\label{tab:3}
\end{center}
\end{table*}

\section{Paraphrase with types and datasets}
Paraphrase identification is a challenging task in many ways. One is the ambiguity of the linguistic definition of paraphrase, which has resulted in a scarcity of manual annotations due to the time-consuming and expert-level linguistic knowledge required. Additionally, there is a lack of comprehensive research on the types of paraphrases present in popular training datasets used in PI and related downstream tasks. As a result, PI is corpus-dependent and struggles to detect sophisticated paraphrases. To address these challenges, we propose a refined typology, \textsc{ReParaphrase}, that is more relevant to computational PI tasks, as well as a BERT-based automatic paraphrase types classifier to assist in augmenting paraphrase datasets.  

\section{BERT-based automatic paraphrase types classifier}

In order to address the issues of ambiguity of linguistic definitions and the scarcity of manual annotation, we propose a refined typology called \textsc{ReParaphrase} and a BERT-based automatic paraphrase types classifier.

The automatic paraphrase types classifier is based on the BERT structure \cite{devlin_bert_2019}. BERT pre-trained models are designed for fine-tuned tasks that use the whole sentence to make decisions, such as sequence classification and token classification. In our case, "BERT base uncased" is chosen for paraphrase type classification. This pre-trained model has 110 million parameters and does not make a distinction between capitalization.

The training dataset is a combination of the Extended Paraphrase Typology Corpus (ETPC) \cite{kovatchev_etpc_2018} and manually annotated datasets. The ETPC dataset was designed for paraphrasing and textual entailment, but it only focuses on a single dataset, leading to unbalanced paraphrase types in the training corpus. Based on the ETPC dataset, we randomly selected paraphrase pairs from six widely used corpora (MSRPC, ParaNMT-50M, PAWS, TaPaCo, ParaBank, WikiQA) and manually annotated them following our \textsc{ReParaphrase} typology. Table \ref{tab:4} in the original text shows some basic information about the training corpus.

\begin{table*}
\centering
\caption{Characteristics of the training corpus}
\begin{tabular}{ p{6cm}  p{7cm}}
\hline
Property &  \\
\hline
Original text units & 2422 \\ 
Text units after adding annotation & 4422\\
Type of text units & sentences pair with '[SEP]' in the middle\\
Words in original corpus & 56783\\
Unique words in original corpus & 6139\\
Max(words) in original pairs & 66\\
Mean(words) in original pairs & 3\\
StDev(words) in original pairs & 11.50\\
Words in new corpus & 177029\\
Unique words in new corpus & 12205\\
Max.(words) in new pairs & 279\\
Mean(words) in new pairs & 3\\
StDev(words) in new pairs & 15.20\\
\hline
\label{tab:4}
\end{tabular}
\end{table*}

We performed our annotation using labelstudio \cite{labelstudio}. We imported selected pairs into label studio and added our typology as instruction. Before annotation, two annotators went through our typology and listed sample sentences for each type (see the section on paraphrase types). They then labeled the same first 200 pairs to assess their agreement on different types. More type details were added to the sample list for annotating the remaining pairs. In the training process, we applied 5-fold Cross-Validation to evaluate our classifier on the annotated datasets. We shuffled the datasets randomly and split them into five groups. We took each unique group as a test data set and the remaining groups as a training data set. We then fit our model on the training set and evaluated it on the test set. Finally, we summarized the performance by taking the average scores. Table \ref{tab:5} shows the performance of a 5-fold evaluation on the last sample of the training corpus.

\begin{table*}
\centering
\caption{Performance of 5-fold evaluation}
\begin{tabular}{ c c c c c}
\hline
Paraphrase type & Precision & Recall & F1-score & Support \\
\hline
Identity&0.99&1.00&0.99&286\\
Inflectional Changes&0.97&0.86&0.91&37\\
Same Polarity Substitution&0.96&0.97&0.96&73\\
Synthetic/Analytic Substitution&0.91&1.00&0.95&52\\
Inflectional Changes &0.87&1.00&0.93&13\\
Change of Order &0.91&0.70&0.79&30\\
Punctuation Changes&0.92&0.65&0.76&17\\
Subordination and Nesting Changes&1.00&1.00&1.00&18\\
Spelling Changes &1.00&0.64&0.78&14\\
Syntax/Discourse Structure Changes&0.67&1.00&0.80&4\\
Change of Format &0.89&0.89&0.89&9\\
Functional Word Substitution &0.86&1.00&0.92&6\\
Ellipsis &1.00&1.00&1.00&2\\
Derivational Changes &0.31&1.00&0.48&5\\
Diathesis Alternation &1.00&0.75&0.86&4\\
Coordination changes&0.86&0.75&0.80&8\\
Direct/indirect Style Alternations&1.00&0.67&0.80&3\\
Non-Paraphrase&0.00&0.00&0.00&2\\
Opposite Polarity Substitution&0.00&0.00&0.00&1\\
Accuracy&    &    &    &0.95\\
Macro avg&0.80&0.78&0.77&884\\
Weighted avg&0.95&0.95&0.95&884\\
\hline
\label{tab:5}
\end{tabular}
\end{table*}
\subsection{Paraphrase corpus}
Our review selects six paraphrase corpora primarily used in PI and related tasks.  
\paragraph{MSRPC}
Microsoft Research Paraphrase Corpus (MSRPC) is a corpus of 5,801 sentence pairs selected from clustering news articles by SVM \cite{dolan_automatically_2005}. Human annotators labeled each pair as a paraphrase or not. Its training subset has 4,076 sentence pairs, of which 2,753 are paraphrases, and the test subset has 1,725 pairs, of which 1,147 are paraphrases.

\paragraph{ParaNMT-50M}
ParaNMT-50M is a dataset of more than 50 million English-English sentential paraphrase pairs. It generated pairs automatically by using neural machine translation to translate the non-English side of a large parallel corpus, following \cite{wieting-etal-2017-learning}.

\paragraph{PAWS}
PAWS is a dataset constructed from Quora and Wikipedia sentences, with 108,463 well-formed paraphrase and non-paraphrase pairs with high lexical overlap. Examples are generated from controlled language models and back translation and given five human ratings each in both phases. A final rule recombines annotated examples and balances the labels\cite{zhang_paws_2019}.

\paragraph{TaPaCo}
Tapaco is a publicly available paraphrase corpus for 73 languages extracted from the Tatoeba database. Tatoeba is a crowdsourcing project mainly geared toward language learners. It aims to provide example sentences and translations for particular linguistic constructions and words. The paraphrase corpus is created by populating a graph with Tatoeba sentences and equivalence links between sentences {``}meaning the same thing{''}. This graph is then traversed to extract sets of paraphrases. Several language-independent filters and pruning steps are applied to remove uninteresting sentences. A manual evaluation performed on three languages shows that between half and three-quarters of inferred paraphrases are correct and that most remaining ones are either correct but trivial or near-paraphrases that neutralize a morphological distinction. The corpus contains a total of 1.9 million sentences, with 200 - 250 000 sentences per language \cite{scherrer_tapaco_2020}.

\paragraph{ParaBank}
Following the approach of ParaNMT\cite{wieting-etal-2017-learning}, ParaBank trains a Czech-English neural machine translation (NMT) system to generate novel paraphrases of English reference sentences. However, by adding lexical constraints to the NMT decoding procedure, they can produce multiple high-quality sentential paraphrases per source sentence, yielding an English paraphrase resource with more than 4 billion generated tokens and exhibiting greater lexical diversity.

\paragraph{WikiQA}
WikiQA is a dataset for open-domain question answering. The dataset contains 3,047 questions originally sampled from Bing query logs. Based on the user clicks, each question is associated with a Wikipedia page presumed to be the topic of the question. In order to eliminate answer sentence biases caused by keyword matching, they consider all the sentences in the summary paragraph of the page as the candidate answer sentences, with labels on whether the sentence is a correct answer to the question provided by crowd-sourcing workers. Among these questions, about one-third of them contain correct answers in the answer sentence set\cite{yang-etal-2015-wikiqa}.

\subsection{Paragraph types distribution of datasets}

The distribution of paraphrase types in existing datasets is unbalanced. Table \ref{tab:6} shows the paraphrase types distribution of MSRPC. "Same Polarity Substitution" and "Identity" are the two top types that occupy 72.6\% of total paraphrase pairs. We have 25 paraphrase types in our annotated corpus, but not all paraphrase types occur in MSRPC. The last 12 types occupy only 4.38\% of total pairs. This unbalanced distribution contributes to the fact that models performing well on MSRPC do not detect all paraphrase types well. We applied our automatic paraphrase types classifier on other datasets and found that unbalanced issues exist in all corpora. Table \ref{tab:7} shows the cumulative percentage of the top 3 types of each dataset. Some auto-generated datasets, such as ParaNMT-50M and Parabank, using back-translation techniques, are more unbalanced in their paraphrase types. Moreover, other manually labeled corpora were used for other paraphrase tasks, which ignore the distribution. As discussed above, different paraphrase types have different features that can be learned as paraphrase patterns for PI. Therefore, an unbalanced distribution leads to insufficient patterns for deep learning models to detect all kinds of paraphrases.
\begin{table}
\centering
\caption{Paraphrase types distribution of MSRPC}
\begin{tabular}{ c  c}
\hline
Type & Counts \\
\hline
Same Polarity Substitution (SPS) &1711\\
Identity (I) &1503\\
Synthetic/Analytic Substitution (S/AS) &263\\
Inflectional changes (IC) &168\\
Change of Order (CO) &152\\
Punctuation Changes (PC) &136\\
Subordination and Nesting Changes (SNC) &97\\
Spelling Changes (SC) &67\\
Syntax/Discourse Structure Changes (S/DSC) &64\\
Change of Format (CF) &62\\
Functional Word Substitution (FWS) &37\\
Ellipsis (Ell) &29\\
Derivational Changes (DC) &28\\
Diathesis Alternation (DA) &27\\
Coordination Changes (CC) &20\\
Converse Substitution (CS) &16\\
Direct/Indirect Style Alternations (D/ISA) &16\\
Addition/Deletion (A/D) &7\\
Sentence Modality Changes (SMC) &5\\
Entailment (Ent) &4\\
Opposite Polarity Substitution (OPS) &3\\
Negation Switching (NS) &2\\
\hline
\label{tab:6}
\end{tabular}
\end{table}

\begin{table*}
\centering
\caption{Paragraph types distribution of datasets}
\begin{tabular}{ c  c c c c} 
\hline
Dataset & Most frequent & Second most frequent & Third most frequent & Accumulated\\
\hline
MSRPC & SPS (38.6\%) & I  (34.0\%) & S/AS (5.9\%) & 78.6\%\\
Parabank & SPS (68.1\%) & I (27.9\%) & PC (1.5\%)& 97.5\%\\
ParaNMT-50M & SPS (90.9\%) & I (1.9\%) & PC (1.7\%) & 94.5\%\\
PAWS & SPS (66.3\%) & SMC (13.8\%) & I (6.6\%) & 86.7\%\\
TaPaCo & SPS (73.7\%) & PC (21.1\%) & I (1.5\%) & 96.3\%\\
WikiQA & I (63.0\%) & PC(28.4\%) & S/DSC (2.7\%) & 94.1\%\\
\hline
\label{tab:7}
\end{tabular}
\end{table*}

\section{New threats posed by Large Language Models: ChatGPT, and other open source LLMs}
Although traditional paraphrase identification approaches provide interpretable and robust solutions to academic or general integrity detection issues, the rapid evolution of Large Language Models (LLMs) raises new threats to paraphrase identification. The new decoding techniques such as top-k and nucleus sampling contribute to more diverse and sophisticated generated sentences compared to previous technologies. For example, prior paraphrase identification approaches mainly focus on document-level or sentence-level features in plagiarism or fake news detection applications. However, LLMs can deceive linguistic pattern detectors with advanced prompts \cite{guo2023close}. Paraphrase identification cannot help with detecting the "hallucination" phenomenon in LLMs which is about generating inconsistent sentences or fake content \cite{zhong-etal-2020-neural}. 

New threats also occur in the data collection process of paraphrase identification tasks. Paraphrase identification models rely on the corpus they are trained to learn textual patterns. However, the biases of data collection can negatively impact the performance and generalization of the models. As we showed in the above experiments, the imbalanced distribution of paraphrase types widely exists in the popular datasets which limits the models' ability to generalize. At the same time, text generated from LLMs is becoming less distinguishable from human-authored ones. As a result, PI-based detection approaches are becoming less feasible in this LLMs era.

Another threat posed by LLMs is confidence calibration \cite{tang2023science}. In the paraphrase identification task, it is crucial not only to have high accuracy of classification but also to get the similarity of paraphrase sentences. Although prior deep learning PI approaches have higher accuracy than the traditional methods, research on accurate confidence scores in text generated from LLMs is scarce. Therefore, the diversity of text produced by LLMs makes it challenging to create similarity measures between paraphrased content.

Easy accessibility of open-source LLMs has led to the degradation of the effectiveness of previous paraphrase identification methods due to LLMs' exceptional ability to generate paraphrases that are more difficult to detect. One solution to detect LLMs generated paraphrase is white-box embedding watermarks which need researchers to have full access to the generative models \cite{kirchenbauer2023reliability}. Open-source LLMs provide an alternative way for users to erase these watermarks, leading to a challenge for paraphrase classification.

In summary, LLMs provide new challenges in solving paraphrase identification, but addressing these threats is crucial to ensuring the effectiveness and reliability of paraphrase identification methods in the era of LLMs.

\section{Limitation of Paraphrase Detection Methods}
While paraphrase identification is based on comparing the semantic information of two documents, methods such as adversarial attacks have been adapted to confuse and disable natural language processing systems for paraphrase identification. For instance, semantic collisions: texts with very different semantic information being predicted as semantically similar by natural language processing models, could be created to attack those systems \cite{song2020adversarial, zhang2019paws}. In \cite{song2020adversarial}, the authors found that word scrambling---swapping words from a sentence, creating sentences with the same lexicon but drastically different semantics---will cause paraphrase identification models to make false positive predictions. While there are many other works proposed for adversarial attacks for breaking NLP systems, it is beyond what we aim to review in this article.

While deep learning and large language models have been used for paraphrase identification tasks, with great performance improvement over the older methods, it could be a concern that those models could be biased and have drastically different performance when serving some groups of users, especially groups that are underrepresented in the training data. While there is no research done related to the social bias in paraphrase detection models, several research studies have found social bias in prominent NLP models such as word embeddings and BERT \cite{10.1145/3351095.3375623, aylin_sem_biases, 10.1145/3461702.3462536}. While great performance could be achieved by using recent pre-trained NLP models, those models could inherit social bias from the base model, giving biased prediction results, or having poor performance on cases related to underrepresented groups.

\section{Discussion}
In this review, we present a comprehensive survey of the current state of paraphrase identification methods and datasets. We focus on both traditional approaches and deep learning-based methods and discuss their strengths and limitations. We also introduce a new typology of paraphrase types and show how different datasets exhibit varying distributions of these types. Our analysis highlights the challenges faced by the paraphrase identification field, particularly the lack of representative paraphrase types in existing training datasets. This calls for future research to develop more robust and scalable methods, as well as larger and more diverse datasets, to advance the state of the art in paraphrase identification.

In this review, we present a new typology of paraphrase types, \textsc{ReParaphrased}, which is more closely aligned with the tasks and challenges of automatic computational paraphrase identification. This typology was refined based on the work of Kovatchev et al. \cite{kovatchev_etpc_2018}, and Bhagat and Hovy \cite{bhagat-hovy-2013-squibs}, and combines some previously overlapping types. By applying this typology to existing datasets, we were able to analyze the distribution of paraphrase types and identify gaps and limitations in the current state of the field. This new typology provides a useful framework for future research on paraphrase identification and may help the community to develop more effective and accurate methods.

According to Table \ref{tab:2}, most deep learning models on PI focus on the compositional distributional semantics of different levels. Structures and techniques are mainly designed to capture the semantic features of paraphrased sentences. Although unsupervised and hybrid learning methods show great performance on PI tasks, supervised learning methods dominated the deep learning methods after 2014 due to the advance of deep neural networks. 

Deep learning approaches offer several key advantages compared to knowledge-based and corpus-based models. These include the ability to capture complex semantic and syntactic features of natural language, handle variable-length sentences, and perform well on tasks that require long-distance dependencies. Despite these advantages, deep learning methods for paraphrase identification also face challenges, such as the need for large amounts of training data and the difficulty of interpretability. Despite these challenges, deep learning continues to be a promising area of research for improving the accuracy and effectiveness of paraphrase identification.

In this review, we surveyed six datasets comprising the majority of studies on paraphrase identification (PI). We randomly sampled pairs from the datasets to create an annotated training dataset. In section 4, we applied our automatic paraphrase types classifier to the six datasets and found similar representational issues of unbalanced types. These issues prompted us to focus on the missing patterns in datasets and other patterns that require more pairs for models to learn. Our review found that some simple paraphrase types, such as "Same Polarity Substitution" and "Identity," often account for a significant proportion of all datasets. Traditional and early deep-learning    approaches can effectively capture these simple paraphrase types. As a result of unbalanced datasets, the performance of some unsupervised learning traditional techniques and early deep learning methods remain competitive, as shown in Table \ref{tab:3}.

PI can be useful for misinformation detection, and correction \cite{parahrase_fake_news, paraphrase_mis_2021}. Since misinformation often shares similar patterns of representation, it can be used as training data for deep learning models. Sophisticated PI techniques for specific domains, such as scientific papers and fake news, can help users track the source of misinformation.

Misinformation is a pervasive problem, and paraphrase identification (PI) can be a valuable tool for detecting and correcting it. By recognizing the similar patterns of representation that misinformation often shares, PI techniques can help deep learning models learn to identify it \cite{parahrase_fake_news, paraphrase_mis_2021}. And by specializing in specific domains, such as scientific papers and fake news, PI can even help users track the source of the misinformation. So if you want to combat the spread of false information, PI might be just the solution you're looking for.

\section{Conclusion}
Paraphrase identification (PI) is an important field of research with many potential applications. In this review, we have examined the most widely used techniques for PI and proposed a new paraphrase type typology. Our typology provides a more comprehensive and nuanced understanding of the different types of paraphrases, which can be useful for the development of new datasets and the training of supervised deep learning models.

We have also highlighted the need for better datasets in order to improve the performance of PI techniques. Current datasets are often unbalanced, with some paraphrase types being overrepresented and others underrepresented. This can lead to errors in classification and hinder the ability of models to learn the patterns of complex paraphrase types. In order to overcome this challenge, we need to produce new datasets that are more balanced and have a larger number of instances.

The field of PI has many potential applications, including the detection of plagiarism and misinformation. In the case of plagiarism, PI techniques can be used to identify instances where the work of others has been inappropriately used without proper attribution. This can be especially important in the education system, where plagiarism can be difficult to detect using traditional methods.

Misinformation is another area where PI can be valuable. By recognizing the similar patterns of representation that misinformation often shares, PI techniques can help deep learning models learn to identify it. And by specializing in specific domains, such as scientific papers and fake news, PI can even help users track the source of the misinformation. This can be a powerful tool for combatting the spread of false information and promoting the dissemination of accurate and reliable information.

In future work, we aim to produce a new dataset based on our proposed paraphrase types and evaluate which models work best with this dataset. We are excited about the potential of PI to improve many aspects of society, and we believe that continued research in this field will lead to significant advances.

\section*{Acknowledgment}
Chao Zhou and Daniel E. Acuna were generously supported by the US's Office of Research Integrity grants ORIIR180041, ORIIIR190049, ORIIIR200052, and ORIIIR210062. Syracuse University's SOURCE grant supported Cheng Qiu.
\bibliographystyle{IEEEtran}
\bibliography{review}

\begin{thebibliography}{100}
\providecommand{\url}[1]{#1}
\csname url@samestyle\endcsname
\providecommand{\newblock}{\relax}
\providecommand{\bibinfo}[2]{#2}
\providecommand{\BIBentrySTDinterwordspacing}{\spaceskip=0pt\relax}
\providecommand{\BIBentryALTinterwordstretchfactor}{4}
\providecommand{\BIBentryALTinterwordspacing}{\spaceskip=\fontdimen2\font plus
\BIBentryALTinterwordstretchfactor\fontdimen3\font minus
  \fontdimen4\font\relax}
\providecommand{\BIBforeignlanguage}[2]{{%
\expandafter\ifx\csname l@#1\endcsname\relax
\typeout{** WARNING: IEEEtran.bst: No hyphenation pattern has been}%
\typeout{** loaded for the language `#1'. Using the pattern for}%
\typeout{** the default language instead.}%
\else
\language=\csname l@#1\endcsname
\fi
#2}}
\providecommand{\BIBdecl}{\relax}
\BIBdecl

\bibitem{brown2020language}
T.~Brown, B.~Mann, N.~Ryder, M.~Subbiah, J.~D. Kaplan, P.~Dhariwal,
  A.~Neelakantan, P.~Shyam, G.~Sastry, A.~Askell \emph{et~al.}, ``Language
  models are few-shot learners,'' \emph{Advances in neural information
  processing systems}, vol.~33, pp. 1877--1901, 2020.

\bibitem{ChatGPT}
\BIBentryALTinterwordspacing
``Chatgpt,'' accessed on December 7, 2022. [Online]. Available:
  \url{https://openai.com/blog/chatgpt/}
\BIBentrySTDinterwordspacing

\bibitem{foltynek_testing_2020}
\BIBentryALTinterwordspacing
T.~Foltynek, D.~Dlabolova, A.~Anohina-Naumeca, S.~Razi, J.~Kravjar, L.~Kamzola,
  J.~Guerrero-Dib, O.~Celik, and D.~Weber-Wulff,
  ``\BIBforeignlanguage{en}{Testing of support tools for plagiarism
  detection},'' \emph{\BIBforeignlanguage{en}{International Journal of
  Educational Technology in Higher Education}}, vol.~17, no.~1, p.~46, Dec.
  2020. [Online]. Available:
  \url{https://educationaltechnologyjournal.springeropen.com/articles/10.118\\6/s41239-020-00192-4}
\BIBentrySTDinterwordspacing

\bibitem{li_simple_2016}
\BIBentryALTinterwordspacing
J.~Li, W.~Monroe, and D.~Jurafsky, ``A {Simple}, {Fast} {Diverse} {Decoding}
  {Algorithm} for {Neural} {Generation},'' Dec. 2016, number: arXiv:1611.08562
  [cs]. [Online]. Available: \url{http://arxiv.org/abs/1611.08562}
\BIBentrySTDinterwordspacing

\bibitem{elder_towards_2018}
\BIBentryALTinterwordspacing
H.~Elder, S.~Gehrmann, A.~O’Connor, and Q.~Lin, ``Towards {Controllable}
  {Generation} of {Diverse} {Natural} {Language},'' in \emph{{INLG} 2018
  ({Challenge} {Track})}, 2018. [Online]. Available:
  \url{https://www.aclweb.org/anthology/W18-6556.pdf}
\BIBentrySTDinterwordspacing

\bibitem{Holtzman2020The}
\BIBentryALTinterwordspacing
A.~Holtzman, J.~Buys, L.~Du, M.~Forbes, and Y.~Choi, ``The curious case of
  neural text degeneration,'' in \emph{International Conference on Learning
  Representations}, 2020. [Online]. Available:
  \url{https://openreview.net/forum?id=rygGQyrFvH}
\BIBentrySTDinterwordspacing

\bibitem{yang_diversity_2021}
S.~Yang, Q.~Zhou, D.~Feng, Y.~Liu, C.~Li, Y.~Cao, and D.~Li, ``Diversity and
  consistency: Exploring visual question-answer pair generation,'' in
  \emph{Findings of the Association for Computational Linguistics: {EMNLP}
  2021}, 2021, pp. 1053--1066.

\bibitem{yang_improving_2022}
E.~Yang, M.~Liu, D.~Xiong, Y.~Zhang, Y.~Meng, J.~Xu, and Y.~Chen, ``Improving
  generation diversity via syntax-controlled paraphrasing,''
  \emph{Neurocomputing}, vol. 485, pp. 103--113, 2022, publisher: Elsevier.

\bibitem{tevet_evaluating_2021}
\BIBentryALTinterwordspacing
G.~Tevet and J.~Berant, ``Evaluating the evaluation of diversity in natural
  language generation,'' 2021, number: {arXiv}:2004.02990. [Online]. Available:
  \url{http://arxiv.org/abs/2004.02990}
\BIBentrySTDinterwordspacing

\bibitem{gupta_abstractive_2019}
\BIBentryALTinterwordspacing
S.~Gupta and S.~K. Gupta, ``Abstractive summarization: An overview of the state
  of the art,'' \emph{Expert Systems with Applications}, vol. 121, pp. 49--65,
  2019. [Online]. Available:
  \url{https://www.sciencedirect.com/science/article/pii/S0957417418307735}
\BIBentrySTDinterwordspacing

\bibitem{clough_developing_2011}
\BIBentryALTinterwordspacing
P.~Clough and M.~Stevenson, ``Developing a corpus of plagiarised short
  answers,'' \emph{Language Resources and Evaluation}, vol.~45, no.~1, pp.
  5--24, 2011. [Online]. Available:
  \url{http://link.springer.com/10.1007/s10579-009-9112-1}
\BIBentrySTDinterwordspacing

\bibitem{nielsen2009recognizing}
R.~D. Nielsen, W.~Ward, and J.~H. Martin, ``Recognizing entailment in
  intelligent tutoring systems,'' \emph{Natural Language Engineering}, vol.~15,
  no.~4, pp. 479--501, 2009.

\bibitem{roe2022automated}
J.~Roe and M.~Perkins, ``What are automated paraphrasing tools and how do we
  address them? a review of a growing threat to academic integrity,''
  \emph{International Journal for Educational Integrity}, vol.~18, no.~1,
  p.~15, 2022.

\bibitem{mckeown_paraphrasing_1983}
K.~{McKeown}, ``Paraphrasing questions using given and new information,''
  \emph{Am. J. Comput. Linguistics}, vol.~9, pp. 1--10, 1983.

\bibitem{hassan-etal-2007-supertagged}
\BIBentryALTinterwordspacing
H.~Hassan, K.~Sima{'}an, and A.~Way, ``Supertagged phrase-based statistical
  machine translation,'' in \emph{Proceedings of the 45th Annual Meeting of the
  Association of Computational Linguistics}.\hskip 1em plus 0.5em minus
  0.4em\relax Prague, Czech Republic: Association for Computational
  Linguistics, Jun. 2007, pp. 288--295. [Online]. Available:
  \url{https://aclanthology.org/P07-1037}
\BIBentrySTDinterwordspacing

\bibitem{bahdanau_neural_2014}
\BIBentryALTinterwordspacing
D.~Bahdanau, K.~Cho, and Y.~Bengio, ``Neural machine translation by jointly
  learning to align and translate,'' \emph{{arXiv}.org perpetual, non-exclusive
  license}, 2014, publisher: {arXiv} Version Number: 7. [Online]. Available:
  \url{https://arxiv.org/abs/1409.0473}
\BIBentrySTDinterwordspacing

\bibitem{cho-etal-2014-learning}
\BIBentryALTinterwordspacing
K.~Cho, B.~van Merri{\"e}nboer, C.~Gulcehre, D.~Bahdanau, F.~Bougares,
  H.~Schwenk, and Y.~Bengio, ``Learning phrase representations using {RNN}
  encoder{--}decoder for statistical machine translation,'' in
  \emph{Proceedings of the 2014 Conference on Empirical Methods in Natural
  Language Processing ({EMNLP})}.\hskip 1em plus 0.5em minus 0.4em\relax Doha,
  Qatar: Association for Computational Linguistics, Oct. 2014, pp. 1724--1734.
  [Online]. Available: \url{https://aclanthology.org/D14-1179}
\BIBentrySTDinterwordspacing

\bibitem{nallapati-etal-2016-abstractive}
\BIBentryALTinterwordspacing
R.~Nallapati, B.~Zhou, C.~dos Santos, C.~Gulcehrer, and B.~Xiang, ``Abstractive
  text summarization using sequence-to-sequence {RNN}s and beyond,'' in
  \emph{Proceedings of The 20th {SIGNLL} Conference on Computational Natural
  Language Learning}.\hskip 1em plus 0.5em minus 0.4em\relax Berlin, Germany:
  Association for Computational Linguistics, Aug. 2016, pp. 280--290. [Online].
  Available: \url{https://aclanthology.org/K16-1028}
\BIBentrySTDinterwordspacing

\bibitem{serban-etal-2017-piecewise}
\BIBentryALTinterwordspacing
I.~V. Serban, A.~G. Ororbia, J.~Pineau, and A.~Courville, ``Piecewise latent
  variables for neural variational text processing,'' in \emph{Proceedings of
  the 2017 Conference on Empirical Methods in Natural Language
  Processing}.\hskip 1em plus 0.5em minus 0.4em\relax Copenhagen, Denmark:
  Association for Computational Linguistics, Sep. 2017, pp. 422--432. [Online].
  Available: \url{https://aclanthology.org/D17-1043}
\BIBentrySTDinterwordspacing

\bibitem{prakash_neural_2016}
\BIBentryALTinterwordspacing
A.~Prakash, S.~A. Hasan, K.~Lee, V.~Datla, A.~Qadir, J.~Liu, and O.~Farri,
  ``Neural paraphrase generation with stacked residual {LSTM} networks,'' in
  \emph{Proceedings of {COLING} 2016, the 26th International Conference on
  Computational Linguistics: Technical Papers}.\hskip 1em plus 0.5em minus
  0.4em\relax The {COLING} 2016 Organizing Committee, 2016, pp. 2923--2934.
  [Online]. Available: \url{https://aclanthology.org/C16-1275}
\BIBentrySTDinterwordspacing

\bibitem{li_paraphrase_2018}
\BIBentryALTinterwordspacing
Z.~Li, X.~Jiang, L.~Shang, and H.~Li, ``Paraphrase generation with deep
  reinforcement learning,'' in \emph{Proceedings of the 2018 Conference on
  Empirical Methods in Natural Language Processing}.\hskip 1em plus 0.5em minus
  0.4em\relax Association for Computational Linguistics, 2018, pp. 3865--3878.
  [Online]. Available: \url{http://aclweb.org/anthology/D18-1421}
\BIBentrySTDinterwordspacing

\bibitem{el2019exploring}
M.~I. El~Desouki and W.~H. Gomaa, ``Exploring the recent trends of paraphrase
  detection,'' \emph{International Journal of Computer Applications}, vol. 975,
  no. S 8887, 2019.

\bibitem{XIAO2020172}
\BIBentryALTinterwordspacing
H.~Xiao, ``Hungarian layer: A novel interpretable neural layer for paraphrase
  identification,'' \emph{Neural Networks}, vol. 131, pp. 172--184, 2020.
  [Online]. Available:
  \url{https://www.sciencedirect.com/science/article/pii/S0893608020302653}
\BIBentrySTDinterwordspacing

\bibitem{hornby_oxford_1995}
\BIBentryALTinterwordspacing
A.~S. Hornby, \emph{Oxford advanced learner's dictionary of current English /
  [by] A.S. Hornby ; editor Jonathan Crowther}.\hskip 1em plus 0.5em minus
  0.4em\relax Fifth edition. Oxford, England : Oxford University Press, 1995.,
  1995. [Online]. Available:
  \url{https://search.library.wisc.edu/catalog/999766027502121}
\BIBentrySTDinterwordspacing

\bibitem{dagan_pascal_2005}
I.~Dagan, O.~Glickman, and B.~Magnini, ``The {PASCAL} recognising textual
  entailment challenge,'' in \emph{MLCW'05: Proceedings of the First
  international conference on Machine Learning Challenges: evaluating
  Predictive Uncertainty Visual Object Classification, and Recognizing Textual
  Entailment}, 2005, pp. 177--190.

\bibitem{dras_tree_nodate}
M.~Dras, ``Tree adjoining grammar and the reluctant paraphrasing of text,''
  \emph{Dras - Tree Adjoining Grammar and the Reluctant
  Paraphras.pdf:files/507/Dras - Tree Adjoining Grammar and the Reluctant
  Paraphras.pdf:application/pdf}, p. 282, 1999.

\bibitem{harris_distributional_1954}
Z.~S. Harris, ``Distributional structure,'' \emph{{WORD}}, vol.~10, no.~2, pp.
  146--162, 1954, publisher: Routledge.

\bibitem{lin_dirt_2001}
\BIBentryALTinterwordspacing
D.~Lin and P.~Pantel, ``{DIRT} @{SBT}@discovery of inference rules from text,''
  in \emph{Proceedings of the seventh {ACM} {SIGKDD} international conference
  on Knowledge discovery and data mining}, ser. {KDD} '01.\hskip 1em plus 0.5em
  minus 0.4em\relax Association for Computing Machinery, 2001, pp. 323--328.
  [Online]. Available: \url{https://doi.org/10.1145/502512.502559}
\BIBentrySTDinterwordspacing

\bibitem{bhagat-hovy-2013-squibs}
\BIBentryALTinterwordspacing
R.~Bhagat and E.~Hovy, ``{S}quibs: What is a paraphrase?'' \emph{Computational
  Linguistics}, vol.~39, no.~3, pp. 463--472, Sep. 2013. [Online]. Available:
  \url{https://aclanthology.org/J13-3001}
\BIBentrySTDinterwordspacing

\bibitem{kovatchev_etpc_2018}
\BIBentryALTinterwordspacing
V.~Kovatchev, M.~A. Martí, and M.~Salamó, ``{ETPC} - a paraphrase
  identification corpus annotated with extended paraphrase typology and
  negation,'' in \emph{Proceedings of the Eleventh International Conference on
  Language Resources and Evaluation ({LREC} 2018)}.\hskip 1em plus 0.5em minus
  0.4em\relax European Language Resources Association ({ELRA}), 2018. [Online].
  Available: \url{https://aclanthology.org/L18-1221}
\BIBentrySTDinterwordspacing

\bibitem{fujita_expanding_2018}
\BIBentryALTinterwordspacing
A.~Fujita and P.~Isabelle, ``Expanding paraphrase lexicons by exploiting
  generalities,'' \emph{{ACM} Transactions on Asian and Low-Resource Language
  Information Processing}, vol.~17, no.~2, pp. 1--36, 2018. [Online].
  Available: \url{https://dl.acm.org/doi/10.1145/3160488}
\BIBentrySTDinterwordspacing

\bibitem{al-smadi_paraphrase_2017}
\BIBentryALTinterwordspacing
M.~{AL}-Smadi, Z.~Jaradat, M.~{AL}-Ayyoub, and Y.~Jararweh, ``Paraphrase
  identification and semantic text similarity analysis in arabic news tweets
  using lexical, syntactic, and semantic features,'' \emph{Information
  Processing and Management: an International Journal}, vol.~53, no.~3, pp.
  640--652, 2017. [Online]. Available:
  \url{https://www.sciencedirect.com/science/article/pii/S0306457316302382}
\BIBentrySTDinterwordspacing

\bibitem{fujita-sato-2008-probabilistic}
\BIBentryALTinterwordspacing
A.~Fujita and S.~Sato, ``A probabilistic model for measuring grammaticality and
  similarity of automatically generated paraphrases of predicate phrases,'' in
  \emph{Proceedings of the 22nd International Conference on Computational
  Linguistics (Coling 2008)}.\hskip 1em plus 0.5em minus 0.4em\relax
  Manchester, UK: Coling 2008 Organizing Committee, Aug. 2008, pp. 225--232.
  [Online]. Available: \url{https://aclanthology.org/C08-1029}
\BIBentrySTDinterwordspacing

\bibitem{miller_wordnet_1995}
\BIBentryALTinterwordspacing
G.~A. Miller, ``{WordNet}: A lexical database for english,'' \emph{Commun.
  {ACM}}, vol.~38, no.~11, pp. 39--41, 1995, place: New York, {NY}, {USA}
  Publisher: Association for Computing Machinery. [Online]. Available:
  \url{https://doi.org/10.1145/219717.219748}
\BIBentrySTDinterwordspacing

\bibitem{mihalcea_corpus-based_2006}
R.~Mihalcea, C.~Corley, and C.~Strapparava, ``Corpus-based and knowledge-based
  measures of text semantic similarity.'' in \emph{Proceedings of the National
  Conference on Artificial Intelligence}, vol.~1, 2006.

\bibitem{finch_using_2005}
\BIBentryALTinterwordspacing
A.~Finch, Y.-S. Hwang, and E.~Sumita, ``Using machine translation evaluation
  techniques to determine sentence-level semantic equivalence,'' in
  \emph{Proceedings of the Third International Workshop on Paraphrasing
  ({IWP}2005)}, 2005. [Online]. Available:
  \url{https://aclanthology.org/I05-5003}
\BIBentrySTDinterwordspacing

\bibitem{cocos_mapping_2017}
\BIBentryALTinterwordspacing
A.~Cocos, M.~Apidianaki, and C.~Callison-Burch, ``Mapping the paraphrase
  database to {WordNet},'' in \emph{Proceedings of the 6th Joint Conference on
  Lexical and Computational Semantics (*{SEM} 2017)}.\hskip 1em plus 0.5em
  minus 0.4em\relax Association for Computational Linguistics, 2017, pp.
  84--90. [Online]. Available: \url{http://www.aclweb.org/anthology/S17-1009}
\BIBentrySTDinterwordspacing

\bibitem{papineni-etal-2002-bleu}
\BIBentryALTinterwordspacing
K.~Papineni, S.~Roukos, T.~Ward, and W.-J. Zhu, ``{B}leu: a method for
  automatic evaluation of machine translation,'' in \emph{Proceedings of the
  40th Annual Meeting of the Association for Computational Linguistics}.\hskip
  1em plus 0.5em minus 0.4em\relax Philadelphia, Pennsylvania, USA: Association
  for Computational Linguistics, Jul. 2002, pp. 311--318. [Online]. Available:
  \url{https://aclanthology.org/P02-1040}
\BIBentrySTDinterwordspacing

\bibitem{10.5555/1289189.1289273}
G.~Doddington, ``Automatic evaluation of machine translation quality using
  n-gram co-occurrence statistics,'' in \emph{Proceedings of the Second
  International Conference on Human Language Technology Research}, ser. HLT
  '02.\hskip 1em plus 0.5em minus 0.4em\relax San Francisco, CA, USA: Morgan
  Kaufmann Publishers Inc., 2002, p. 138–145.

\bibitem{WER-1992}
\BIBentryALTinterwordspacing
K.-Y. Su, M.-W. Wu, and J.-S. Chang, ``A new quantitative quality measure for
  machine translation systems,'' in \emph{Proceedings of the 14th Conference on
  Computational Linguistics - Volume 2}, ser. COLING '92.\hskip 1em plus 0.5em
  minus 0.4em\relax USA: Association for Computational Linguistics, 1992, p.
  433–439. [Online]. Available: \url{https://doi.org/10.3115/992133.992137}
\BIBentrySTDinterwordspacing

\bibitem{Tillmann97accelerateddp}
C.~Tillmann, S.~Vogel, H.~Ney, A.~Zubiaga, and H.~Sawaf, ``Accelerated dp based
  search for statistical translation,'' in \emph{In European Conf. on Speech
  Communication and Technology}, 1997, pp. 2667--2670.

\bibitem{wan_using_2006}
\BIBentryALTinterwordspacing
S.~Wan, M.~Dras, R.~Dale, and C.~Paris, ``Using dependency-based features to
  take the 'para-farce' out of paraphrase,'' in \emph{Proceedings of the
  Australasian Language Technology Workshop 2006}, 2006, pp. 131--138.
  [Online]. Available: \url{https://aclanthology.org/U06-1019}
\BIBentrySTDinterwordspacing

\bibitem{madnani_re-examining_2012}
\BIBentryALTinterwordspacing
N.~Madnani, J.~Tetreault, and M.~Chodorow, ``Re-examining machine translation
  metrics for paraphrase identification,'' in \emph{Proceedings of the 2012
  Conference of the North American Chapter of the Association for Computational
  Linguistics: Human Language Technologies}.\hskip 1em plus 0.5em minus
  0.4em\relax Association for Computational Linguistics, 2012, pp. 182--190.
  [Online]. Available: \url{https://aclanthology.org/N12-1019}
\BIBentrySTDinterwordspacing

\bibitem{10.2307/40783463}
\BIBentryALTinterwordspacing
M.~G. Snover, N.~Madnani, B.~Dorr, and R.~Schwartz, ``Ter-plus: paraphrase,
  semantic, and alignment enhancements to translation edit rate,''
  \emph{Machine Translation}, vol.~23, no. 2/3, pp. 117--127, 2009. [Online].
  Available: \url{http://www.jstor.org/stable/40783463}
\BIBentrySTDinterwordspacing

\bibitem{denkowski-lavie-2010-extending}
\BIBentryALTinterwordspacing
M.~Denkowski and A.~Lavie, ``Extending the {METEOR} machine translation
  evaluation metric to the phrase level,'' in \emph{Human Language
  Technologies: The 2010 Annual Conference of the North {A}merican Chapter of
  the Association for Computational Linguistics}.\hskip 1em plus 0.5em minus
  0.4em\relax Los Angeles, California: Association for Computational
  Linguistics, Jun. 2010, pp. 250--253. [Online]. Available:
  \url{https://aclanthology.org/N10-1031}
\BIBentrySTDinterwordspacing

\bibitem{habash2008sepia}
N.~Habash and A.~Elkholy, ``Sepia: surface span extension to syntactic
  dependency precision-based mt evaluation,'' in \emph{Proceedings of the NIST
  metrics for machine translation workshop at the association for machine
  translation in the Americas conference, AMTA-2008. Waikiki, HI}.\hskip 1em
  plus 0.5em minus 0.4em\relax Citeseer, 2008.

\bibitem{Parker_badger:a}
S.~Parker, ``Badger: A new machine translation metric,'' 2008.

\bibitem{chan-ng-2008-maxsim}
\BIBentryALTinterwordspacing
Y.~S. Chan and H.~T. Ng, ``{MAXSIM}: A maximum similarity metric for machine
  translation evaluation,'' in \emph{Proceedings of ACL-08: HLT}.\hskip 1em
  plus 0.5em minus 0.4em\relax Columbus, Ohio: Association for Computational
  Linguistics, Jun. 2008, pp. 55--62. [Online]. Available:
  \url{https://aclanthology.org/P08-1007}
\BIBentrySTDinterwordspacing

\bibitem{potthast2009pan}
M.~Potthast, A.~Eiselt, B.~Stein, A.~Barr{\'o}nCedeno, and P.~Rosso, ``Pan
  plagiarism corpus pan-pc-09,'' \emph{Online: http://www. uniweimar.
  de/cms/medien/webis/research/corpora/pan-pc-09. html--Date accessed}, vol.~2,
  no.~03, p. 2010, 2009.

\bibitem{taylor_penn_2003}
A.~Taylor, M.~Marcus, and B.~Santorini, ``The penn treebank: An overview,''
  2003.

\bibitem{das_paraphrase_2009}
\BIBentryALTinterwordspacing
D.~Das and N.~A. Smith, ``Paraphrase identification as probabilistic
  quasi-synchronous recognition,'' in \emph{Proceedings of the Joint Conference
  of the 47th Annual Meeting of the {ACL} and the 4th International Joint
  Conference on Natural Language Processing of the {AFNLP}}.\hskip 1em plus
  0.5em minus 0.4em\relax Association for Computational Linguistics, 2009, pp.
  468--476. [Online]. Available: \url{https://aclanthology.org/P09-1053}
\BIBentrySTDinterwordspacing

\bibitem{sidorov_syntactic_2013}
G.~Sidorov, ``Syntactic dependency based n-grams in rule based automatic
  english as second language grammar correction,'' \emph{Int. J. Comput.
  Linguistics Appl.}, vol.~4, pp. 169--188, 2013.

\bibitem{wong-mooney-2006-learning}
\BIBentryALTinterwordspacing
Y.~W. Wong and R.~Mooney, ``Learning for semantic parsing with statistical
  machine translation,'' in \emph{Proceedings of the Human Language Technology
  Conference of the {NAACL}, Main Conference}.\hskip 1em plus 0.5em minus
  0.4em\relax New York City, USA: Association for Computational Linguistics,
  Jun. 2006, pp. 439--446. [Online]. Available:
  \url{https://aclanthology.org/N06-1056}
\BIBentrySTDinterwordspacing

\bibitem{zettlemoyer-collins-2009-learning}
\BIBentryALTinterwordspacing
L.~Zettlemoyer and M.~Collins, ``Learning context-dependent mappings from
  sentences to logical form,'' in \emph{Proceedings of the Joint Conference of
  the 47th Annual Meeting of the {ACL} and the 4th International Joint
  Conference on Natural Language Processing of the {AFNLP}}.\hskip 1em plus
  0.5em minus 0.4em\relax Suntec, Singapore: Association for Computational
  Linguistics, Aug. 2009, pp. 976--984. [Online]. Available:
  \url{https://aclanthology.org/P09-1110}
\BIBentrySTDinterwordspacing

\bibitem{sun2014empirical}
X.~Sun, X.~Liu, J.~Hu, and J.~Zhu, ``Empirical studies on the nlp techniques
  for source code data preprocessing,'' in \emph{Proceedings of the 2014 3rd
  international workshop on evidential assessment of software technologies},
  2014, pp. 32--39.

\bibitem{bao_knowledge-based_2014}
\BIBentryALTinterwordspacing
J.~Bao, N.~Duan, M.~Zhou, and T.~Zhao, ``Knowledge-based question answering as
  machine translation,'' in \emph{Proceedings of the 52nd Annual Meeting of the
  Association for Computational Linguistics (Volume 1: Long Papers)}.\hskip 1em
  plus 0.5em minus 0.4em\relax Association for Computational Linguistics, 2014,
  pp. 967--976. [Online]. Available: \url{http://aclweb.org/anthology/P14-1091}
\BIBentrySTDinterwordspacing

\bibitem{dong_statistical_2015}
\BIBentryALTinterwordspacing
L.~Dong, F.~Wei, S.~Liu, M.~Zhou, and K.~Xu, ``A statistical parsing framework
  for sentiment classification,'' \emph{Computational Linguistics}, vol.~41,
  no.~2, pp. 293--336, 2015. [Online]. Available:
  \url{https://direct.mit.edu/coli/article/41/2/293-336/1511}
\BIBentrySTDinterwordspacing

\bibitem{bouamor_multitechnique_2013}
\BIBentryALTinterwordspacing
H.~Bouamor, A.~Max, and A.~Vilnat, ``Multitechnique paraphrase alignment: A
  contribution to pinpointing sub-sentential paraphrases,'' \emph{{ACM}
  Transactions on Intelligent Systems and Technology}, vol.~4, no.~3, pp.
  1--27, 2013. [Online]. Available:
  \url{https://dl.acm.org/doi/10.1145/2483669.2483677}
\BIBentrySTDinterwordspacing

\bibitem{flanigan_discriminative_2014}
\BIBentryALTinterwordspacing
J.~Flanigan, S.~Thomson, J.~Carbonell, C.~Dyer, and N.~A. Smith, ``A
  discriminative graph-based parser for the abstract meaning representation,''
  in \emph{Proceedings of the 52nd Annual Meeting of the Association for
  Computational Linguistics (Volume 1: Long Papers)}.\hskip 1em plus 0.5em
  minus 0.4em\relax Association for Computational Linguistics, 2014, pp.
  1426--1436. [Online]. Available: \url{http://aclweb.org/anthology/P14-1134}
\BIBentrySTDinterwordspacing

\bibitem{peng_synchronous_2015}
\BIBentryALTinterwordspacing
X.~Peng, L.~Song, and D.~Gildea, ``A synchronous hyperedge replacement grammar
  based approach for {AMR} parsing,'' in \emph{Proceedings of the Nineteenth
  Conference on Computational Natural Language Learning}.\hskip 1em plus 0.5em
  minus 0.4em\relax Association for Computational Linguistics, 2015, pp.
  32--41. [Online]. Available: \url{http://aclweb.org/anthology/K15-1004}
\BIBentrySTDinterwordspacing

\bibitem{zhou_amr_2016}
\BIBentryALTinterwordspacing
J.~Zhou, F.~Xu, H.~Uszkoreit, W.~Qu, R.~Li, and Y.~Gu, ``{AMR} parsing with an
  incremental joint model,'' in \emph{Proceedings of the 2016 Conference on
  Empirical Methods in Natural Language Processing}.\hskip 1em plus 0.5em minus
  0.4em\relax Association for Computational Linguistics, 2016, pp. 680--689.
  [Online]. Available: \url{http://aclweb.org/anthology/D16-1065}
\BIBentrySTDinterwordspacing

\bibitem{konstas_neural_2017}
\BIBentryALTinterwordspacing
I.~Konstas, S.~Iyer, M.~Yatskar, Y.~Choi, and L.~Zettlemoyer, ``Neural {AMR}:
  Sequence-to-sequence models for parsing and generation,'' 2017. [Online].
  Available: \url{http://arxiv.org/abs/1704.08381}
\BIBentrySTDinterwordspacing

\bibitem{issa_abstract_2018}
\BIBentryALTinterwordspacing
F.~Issa, M.~Damonte, S.~B. Cohen, X.~Yan, and Y.~Chang, ``Abstract meaning
  representation for paraphrase detection,'' in \emph{Proceedings of the 2018
  Conference of the North American Chapter of the Association for Computational
  Linguistics: Human Language Technologies, Volume 1 (Long Papers)}.\hskip 1em
  plus 0.5em minus 0.4em\relax Association for Computational Linguistics, 2018,
  pp. 442--452. [Online]. Available: \url{http://aclweb.org/anthology/N18-1041}
\BIBentrySTDinterwordspacing

\bibitem{blloshmi_xl-amr_2020}
\BIBentryALTinterwordspacing
R.~Blloshmi, R.~Tripodi, and R.~Navigli, ``{XL}-{AMR}: Enabling cross-lingual
  {AMR} parsing with transfer learning techniques,'' in \emph{Proceedings of
  the 2020 Conference on Empirical Methods in Natural Language Processing
  ({EMNLP})}.\hskip 1em plus 0.5em minus 0.4em\relax Association for
  Computational Linguistics, 2020, pp. 2487--2500. [Online]. Available:
  \url{https://www.aclweb.org/anthology/2020.emnlp-main.195}
\BIBentrySTDinterwordspacing

\bibitem{ribeiro_structural_2021}
\BIBentryALTinterwordspacing
L.~F.~R. Ribeiro, Y.~Zhang, and I.~Gurevych, ``Structural adapters in
  pretrained language models for {AMR}-to-text generation,'' 2021. [Online].
  Available: \url{http://arxiv.org/abs/2103.09120}
\BIBentrySTDinterwordspacing

\bibitem{doi:10.1080/01638539809545028}
\BIBentryALTinterwordspacing
T.~K. Landauer, P.~W. Foltz, and D.~Laham, ``An introduction to latent semantic
  analysis,'' \emph{Discourse Processes}, vol.~25, no. 2-3, pp. 259--284, 1998.
  [Online]. Available: \url{https://doi.org/10.1080/01638539809545028}
\BIBentrySTDinterwordspacing

\bibitem{guo-diab-2012-modeling}
\BIBentryALTinterwordspacing
W.~Guo and M.~Diab, ``Modeling sentences in the latent space,'' in
  \emph{Proceedings of the 50th Annual Meeting of the Association for
  Computational Linguistics (Volume 1: Long Papers)}.\hskip 1em plus 0.5em
  minus 0.4em\relax Jeju Island, Korea: Association for Computational
  Linguistics, Jul. 2012, pp. 864--872. [Online]. Available:
  \url{https://aclanthology.org/P12-1091}
\BIBentrySTDinterwordspacing

\bibitem{ji_discriminative_2013}
Y.~Ji and J.~Eisenstein, ``Discriminative improvements to distributional
  sentence similarity,'' in \emph{{EMNLP}}, 2013.

\bibitem{10.5555/3008751.3008829}
D.~D. Lee and H.~S. Seung, ``Algorithms for non-negative matrix
  factorization,'' in \emph{Proceedings of the 13th International Conference on
  Neural Information Processing Systems}, ser. NIPS'00.\hskip 1em plus 0.5em
  minus 0.4em\relax Cambridge, MA, USA: MIT Press, 2000, p. 535–541.

\bibitem{yin_discriminative_2016}
\BIBentryALTinterwordspacing
W.~Yin and H.~Schütze, ``Discriminative phrase embedding for paraphrase
  identification,'' 2016. [Online]. Available:
  \url{http://arxiv.org/abs/1604.00503}
\BIBentrySTDinterwordspacing

\bibitem{Mar13}
A.~A. Markov, ``Essai d'une recherche statistique sur le texte du roman
  ``{E}ugene {O}negin'' illustrant la liaison des epreuve en chain ({`Example}
  of a statistical investigation of the text of ``{E}ugene {O}negin"
  illustrating the dependence between samples in chain'),'' \emph{Izvistia
  Imperatorskoi Akademii Nauk (Bulletin de {l'Acad\'{e}mie} {Imp\'{e}riale} des
  Sciences de {St.-P\'{e}tersbourg)}}, vol.~7, pp. 153--162, 1913, english
  translation by Morris Halle, 1956.

\bibitem{shannon1948mathematical}
C.~E. Shannon, ``A mathematical theory of communication,'' \emph{The Bell
  system technical journal}, vol.~27, no.~3, pp. 379--423, 1948.

\bibitem{1454428}
F.~Jelinek, ``Continuous speech recognition by statistical methods,''
  \emph{Proceedings of the IEEE}, vol.~64, no.~4, pp. 532--556, 1976.

\bibitem{gavalda-waibel-1998-growing-semantic}
\BIBentryALTinterwordspacing
M.~Gavalda and A.~Waibel, ``Growing semantic grammars,'' in \emph{36th Annual
  Meeting of the Association for Computational Linguistics and 17th
  International Conference on Computational Linguistics, Volume 1}.\hskip 1em
  plus 0.5em minus 0.4em\relax Montreal, Quebec, Canada: Association for
  Computational Linguistics, Aug. 1998, pp. 451--456. [Online]. Available:
  \url{https://aclanthology.org/P98-1075}
\BIBentrySTDinterwordspacing

\bibitem{smith-eisner-2006-quasi}
\BIBentryALTinterwordspacing
D.~Smith and J.~Eisner, ``Quasi-synchronous grammars: Alignment by soft
  projection of syntactic dependencies,'' in \emph{Proceedings on the Workshop
  on Statistical Machine Translation}.\hskip 1em plus 0.5em minus 0.4em\relax
  New York City: Association for Computational Linguistics, Jun. 2006, pp.
  23--30. [Online]. Available: \url{https://aclanthology.org/W06-3104}
\BIBentrySTDinterwordspacing

\bibitem{wang2007jeopardy}
M.~Wang, N.~A. Smith, and T.~Mitamura, ``What is the jeopardy model? a
  quasi-synchronous grammar for qa,'' in \emph{Proceedings of the 2007 joint
  conference on empirical methods in natural language processing and
  computational natural language learning (EMNLP-CoNLL)}, 2007, pp. 22--32.

\bibitem{sidorov_syntactic_2014}
\BIBentryALTinterwordspacing
G.~Sidorov, F.~Velasquez, E.~Stamatatos, A.~Gelbukh, and L.~Chanona-Hernández,
  ``Syntactic n-grams as machine learning features for natural language
  processing,'' \emph{Expert Systems with Applications}, vol.~41, no.~3, pp.
  853--860, 2014. [Online]. Available:
  \url{https://www.sciencedirect.com/science/article/pii/S0957417413006271}
\BIBentrySTDinterwordspacing

\bibitem{calvo_dependency_2014}
\BIBentryALTinterwordspacing
H.~Calvo, A.~Segura-Olivares, and A.~García, ``Dependency vs. constituent
  based syntactic n-grams in text similarity measures for paraphrase
  recognition,'' \emph{Computación y Sistemas}, vol.~18, no.~3, 2014.
  [Online]. Available:
  \url{http://cys.cic.ipn.mx/ojs/index.php/CyS/article/view/2044}
\BIBentrySTDinterwordspacing

\bibitem{10.5555/1214993}
D.~Jurafsky and J.~H. Martin, \emph{Speech and Language Processing (2nd
  Edition)}.\hskip 1em plus 0.5em minus 0.4em\relax USA: Prentice-Hall, Inc.,
  2009.

\bibitem{brockett_support_2005}
\BIBentryALTinterwordspacing
C.~Brockett and W.~B. Dolan, ``Support vector machines for paraphrase
  identification and corpus construction,'' in \emph{Proceedings of the Third
  International Workshop on Paraphrasing ({IWP}2005)}, 2005. [Online].
  Available: \url{https://aclanthology.org/I05-5001}
\BIBentrySTDinterwordspacing

\bibitem{dolan_unsupervised_2004}
\BIBentryALTinterwordspacing
B.~Dolan, C.~Quirk, and C.~Brockett, ``Unsupervised construction of large
  paraphrase corpora: Exploiting massively parallel news sources,'' in
  \emph{{COLING} 2004: Proceedings of the 20th International Conference on
  Computational Linguistics}.\hskip 1em plus 0.5em minus 0.4em\relax {COLING},
  2004, pp. 350--356. [Online]. Available:
  \url{https://aclanthology.org/C04-1051}
\BIBentrySTDinterwordspacing

\bibitem{park_sentential_2014}
\BIBentryALTinterwordspacing
H.~Park, G.~Gweon, H.-J. Choi, J.~Heo, and P.-M. Ryu, ``Sentential paraphrase
  generation for agglutinative languages using {SVM} with a string kernel,'' in
  \emph{Proceedings of the 28th Pacific Asia Conference on Language,
  Information and Computing}.\hskip 1em plus 0.5em minus 0.4em\relax Department
  of Linguistics, Chulalongkorn University, 2014, pp. 650--657. [Online].
  Available: \url{https://aclanthology.org/Y14-1074}
\BIBentrySTDinterwordspacing

\bibitem{CC01a}
C.-C. Chang and C.-J. Lin, ``{LIBSVM}: A library for support vector machines,''
  \emph{ACM Transactions on Intelligent Systems and Technology}, vol.~2, pp.
  27:1--27:27, 2011, software available at
  \url{http://www.csie.ntu.edu.tw/~cjlin/libsvm}.

\bibitem{eyecioglu_twitter_2015}
\BIBentryALTinterwordspacing
A.~Eyecioglu and B.~Keller, ``Twitter paraphrase identification with simple
  overlap features and {SVMs},'' in \emph{Proceedings of the 9th International
  Workshop on Semantic Evaluation ({SemEval} 2015)}.\hskip 1em plus 0.5em minus
  0.4em\relax Association for Computational Linguistics, 2015, pp. 64--69.
  [Online]. Available: \url{http://aclweb.org/anthology/S15-2011}
\BIBentrySTDinterwordspacing

\bibitem{rumelhart_learning_1986}
\BIBentryALTinterwordspacing
D.~E. Rumelhart, G.~E. Hinton, and R.~J. Williams, ``Learning representations
  by back-propagating errors,'' \emph{Nature}, vol. 323, no. 6088, pp.
  533--536, 1986. [Online]. Available: \url{https://doi.org/10.1038/323533a0}
\BIBentrySTDinterwordspacing

\bibitem{Deerwester1990IndexingBL}
S.~C. Deerwester, S.~T. Dumais, T.~K. Landauer, G.~W. Furnas, and R.~A.
  Harshman, ``Indexing by latent semantic analysis,'' \emph{J. Am. Soc. Inf.
  Sci.}, vol.~41, pp. 391--407, 1990.

\bibitem{Neural-Pro-LM-2003}
Y.~Bengio, R.~Ducharme, P.~Vincent, and C.~Janvin, ``A neural probabilistic
  language model,'' \emph{J. Mach. Learn. Res.}, vol.~3, no. null, p.
  1137–1155, 3 2003.

\bibitem{doi:10.1126/science.1127647}
\BIBentryALTinterwordspacing
G.~E. Hinton and R.~R. Salakhutdinov, ``Reducing the dimensionality of data
  with neural networks,'' \emph{Science}, vol. 313, no. 5786, pp. 504--507,
  2006. [Online]. Available:
  \url{https://www.science.org/doi/abs/10.1126/science.1127647}
\BIBentrySTDinterwordspacing

\bibitem{bengio_greedy_2006}
Y.~Bengio, P.~Lamblin, D.~Popovici, and H.~Larochelle, ``Greedy layer-wise
  training of deep networks,'' in \emph{Advances in Neural Information
  Processing Systems}, B.~Schölkopf, J.~Platt, and T.~Hoffman, Eds.,
  vol.~19.\hskip 1em plus 0.5em minus 0.4em\relax {MIT} Press, 2006.

\bibitem{Kombrink_Stefan_Mikolov_2011}
S.~Kombrink, T.~Mikolov, M.~Karafiát, and L.~Burget, ``Recurrent neural
  network based language modeling in meeting recognition.'' in
  \emph{Proceedings of the Annual Conference of the International Speech
  Communication Association, INTERSPEECH}, 01 2011, pp. 2877--2880.

\bibitem{blacoe-lapata-2012-comparison}
\BIBentryALTinterwordspacing
W.~Blacoe and M.~Lapata, ``A comparison of vector-based representations for
  semantic composition,'' in \emph{Proceedings of the 2012 Joint Conference on
  Empirical Methods in Natural Language Processing and Computational Natural
  Language Learning}.\hskip 1em plus 0.5em minus 0.4em\relax Jeju Island,
  Korea: Association for Computational Linguistics, Jul. 2012, pp. 546--556.
  [Online]. Available: \url{https://aclanthology.org/D12-1050}
\BIBentrySTDinterwordspacing

\bibitem{banea_simcompass_2014}
\BIBentryALTinterwordspacing
C.~Banea, D.~Chen, R.~Mihalcea, C.~Cardie, and J.~Wiebe, ``{SimCompass}: Using
  deep learning word embeddings to assess cross-level similarity,'' in
  \emph{Proceedings of the 8th International Workshop on Semantic Evaluation
  ({SemEval} 2014)}.\hskip 1em plus 0.5em minus 0.4em\relax Association for
  Computational Linguistics, 2014, pp. 560--565. [Online]. Available:
  \url{http://aclweb.org/anthology/S14-2098}
\BIBentrySTDinterwordspacing

\bibitem{yu2014deep}
L.~Yu, K.~M. Hermann, P.~Blunsom, and S.~Pulman, ``Deep learning for answer
  sentence selection,'' \emph{arXiv preprint arXiv:1412.1632}, 2014.

\bibitem{he_pairwise_2016}
\BIBentryALTinterwordspacing
H.~He and J.~Lin, ``Pairwise word interaction modeling with deep neural
  networks for semantic similarity measurement,'' in \emph{Proceedings of the
  2016 Conference of the North American Chapter of the Association for
  Computational Linguistics: Human Language Technologies}.\hskip 1em plus 0.5em
  minus 0.4em\relax Association for Computational Linguistics, 2016, pp.
  937--948. [Online]. Available: \url{http://aclweb.org/anthology/N16-1108}
\BIBentrySTDinterwordspacing

\bibitem{rocktaschel2016reasoning}
T.~Rocktaschel, E.~Grefenstette, K.~M. Hermann, T.~Kocisky, and P.~Blunsom,
  ``Reasoning about entailment with neural attention,'' in \emph{International
  Conference on Learning Representations (ICLR)}, 2 2016.

\bibitem{shen_baseline_2018}
\BIBentryALTinterwordspacing
D.~Shen, G.~Wang, W.~Wang, M.~R. Min, Q.~Su, Y.~Zhang, C.~Li, R.~Henao, and
  L.~Carin, ``Baseline needs more love: On simple word-embedding-based models
  and associated pooling mechanisms,'' in \emph{Proceedings of the 56th Annual
  Meeting of the Association for Computational Linguistics (Volume 1: Long
  Papers)}.\hskip 1em plus 0.5em minus 0.4em\relax Association for
  Computational Linguistics, 2018, pp. 440--450. [Online]. Available:
  \url{http://aclweb.org/anthology/P18-1041}
\BIBentrySTDinterwordspacing

\bibitem{tan2018multiway}
C.~Tan, F.~Wei, W.~Wang, W.~Lv, and M.~Zhou, ``Multiway attention networks for
  modeling sentence pairs.'' in \emph{IJCAI}, 2018, pp. 4411--4417.

\bibitem{socher_dynamic_2011}
\BIBentryALTinterwordspacing
R.~Socher, E.~Huang, J.~Pennin, C.~D. Manning, and A.~Ng, ``Dynamic pooling and
  unfolding recursive autoencoders for paraphrase detection,'' in
  \emph{Advances in Neural Information Processing Systems}, J.~Shawe-Taylor,
  R.~Zemel, P.~Bartlett, F.~Pereira, and K.~Q. Weinberger, Eds., vol.~24.\hskip
  1em plus 0.5em minus 0.4em\relax Curran Associates, Inc., 2011. [Online].
  Available:
  \url{https://proceedings.neurips.cc/paper/2011/file/3335881e06d4d2309138\\9226225e17c7-Paper.pdf}
\BIBentrySTDinterwordspacing

\bibitem{huang2011paraphrase}
E.~Huang, ``Paraphrase detection using recursive autoencoder,''
  \emph{Source:[http://nlp. stanford. edu/courses/cs224n/2011/reports/ehhuang.
  pdf]}, 2011.

\bibitem{DBLP:journals/corr/ChoMGBSB14}
\BIBentryALTinterwordspacing
K.~Cho, B.~van Merrienboer, {\c{C}}.~G{\"{u}}l{\c{c}}ehre, F.~Bougares,
  H.~Schwenk, and Y.~Bengio, ``Learning phrase representations using {RNN}
  encoder-decoder for statistical machine translation,'' \emph{CoRR}, 2014.
  [Online]. Available: \url{http://arxiv.org/abs/1406.1078}
\BIBentrySTDinterwordspacing

\bibitem{Synactic-tree-LSTM-2015}
\BIBentryALTinterwordspacing
K.~S. Tai, R.~Socher, and C.~D. Manning, ``Improved semantic representations
  from tree-structured long short-term memory networks,'' \emph{CoRR}, vol.
  abs/1503.00075, 2015. [Online]. Available:
  \url{http://arxiv.org/abs/1503.00075}
\BIBentrySTDinterwordspacing

\bibitem{peng2022towards}
Q.~Peng, D.~Weir, and J.~Weeds, ``Towards structure-aware paraphrase
  identification with phrase alignment using sentence encoders,'' \emph{arXiv
  preprint arXiv:2210.05302}, 2022.

\bibitem{DBLP:journals/corr/KirosZSZTUF15}
\BIBentryALTinterwordspacing
R.~Kiros, Y.~Zhu, R.~Salakhutdinov, R.~S. Zemel, A.~Torralba, R.~Urtasun, and
  S.~Fidler, ``Skip-thought vectors,'' \emph{CoRR}, vol. abs/1506.06726, 2015.
  [Online]. Available: \url{http://arxiv.org/abs/1506.06726}
\BIBentrySTDinterwordspacing

\bibitem{Hu-et-al-2014}
B.~Hu, Z.~Lu, H.~Li, and Q.~Chen, ``Convolutional neural network architectures
  for matching natural language sentences,'' in \emph{NIPS}, ser.
  NIPS'14.\hskip 1em plus 0.5em minus 0.4em\relax Cambridge, MA, USA: MIT
  Press, 2014, p. 2042–2050.

\bibitem{ABCNN-yin-2015}
\BIBentryALTinterwordspacing
W.~Yin, H.~Sch{\"{u}}tze, B.~Xiang, and B.~Zhou, ``{ABCNN:} attention-based
  convolutional neural network for modeling sentence pairs,'' \emph{CoRR}, vol.
  abs/1512.05193, 2015. [Online]. Available:
  \url{http://arxiv.org/abs/1512.05193}
\BIBentrySTDinterwordspacing

\bibitem{Multipossen2016}
S.~Wan, Y.~Lan, J.~Guo, J.~Xu, L.~Pang, and X.~Cheng, ``A deep architecture for
  semantic matching with multiple positional sentence representations,'' in
  \emph{Proceedings of the Thirtieth AAAI Conference on Artificial
  Intelligence}, ser. AAAI'16.\hskip 1em plus 0.5em minus 0.4em\relax AAAI
  Press, 2016, p. 2835–2841.

\bibitem{tay2018co}
Y.~Tay, L.~A. Tuan, and S.~C. Hui, ``Co-stack residual affinity networks with
  multi-level attention refinement for matching text sequences,'' \emph{arXiv
  preprint arXiv:1810.02938}, 2018.

\bibitem{Collobert_and_Weston_2008}
\BIBentryALTinterwordspacing
R.~Collobert and J.~Weston, ``A unified architecture for natural language
  processing: Deep neural networks with multitask learning,'' in
  \emph{Proceedings of the 25th International Conference on Machine Learning},
  ser. ICML '08.\hskip 1em plus 0.5em minus 0.4em\relax New York, NY, USA:
  Association for Computing Machinery, 2008, p. 160–167. [Online]. Available:
  \url{https://doi.org/10.1145/1390156.1390177}
\BIBentrySTDinterwordspacing

\bibitem{yin_convolutional_2015}
\BIBentryALTinterwordspacing
W.~Yin and H.~Schütze, ``Convolutional neural network for paraphrase
  identification,'' in \emph{Proceedings of the 2015 Conference of the North
  American Chapter of the Association for Computational Linguistics: Human
  Language Technologies}.\hskip 1em plus 0.5em minus 0.4em\relax Association
  for Computational Linguistics, 2015, pp. 901--911. [Online]. Available:
  \url{http://aclweb.org/anthology/N15-1091}
\BIBentrySTDinterwordspacing

\bibitem{DiSAN-Tao-2017}
\BIBentryALTinterwordspacing
T.~Shen, T.~Zhou, G.~Long, J.~Jiang, S.~Pan, and C.~Zhang, ``Disan: Directional
  self-attention network for rnn/cnn-free language understanding,''
  \emph{CoRR}, vol. abs/1709.04696, 2017. [Online]. Available:
  \url{http://arxiv.org/abs/1709.04696}
\BIBentrySTDinterwordspacing

\bibitem{he-etal-2015-multi}
\BIBentryALTinterwordspacing
H.~He, K.~Gimpel, and J.~Lin, ``Multi-perspective sentence similarity modeling
  with convolutional neural networks,'' in \emph{Proceedings of the 2015
  Conference on Empirical Methods in Natural Language Processing}.\hskip 1em
  plus 0.5em minus 0.4em\relax Lisbon, Portugal: Association for Computational
  Linguistics, Sep. 2015, pp. 1576--1586. [Online]. Available:
  \url{https://aclanthology.org/D15-1181}
\BIBentrySTDinterwordspacing

\bibitem{wang2017bilateral}
Z.~Wang, W.~Hamza, and R.~Florian, ``Bilateral multi-perspective matching for
  natural language sentences,'' \emph{arXiv preprint}, 2017.

\bibitem{gong2017natural}
Y.~Gong, H.~Luo, and J.~Zhang, ``Natural language inference over interaction
  space,'' \emph{arXiv preprint arXiv:1709.04348}, 2017.

\bibitem{ARASE2021101164}
\BIBentryALTinterwordspacing
Y.~Arase and J.~Tsujii, ``Transfer fine-tuning of bert with phrasal
  paraphrases,'' \emph{Computer Speech \& Language}, vol.~66, p. 101164, 2021.
  [Online]. Available:
  \url{https://www.sciencedirect.com/science/article/pii/S0885230820300978}
\BIBentrySTDinterwordspacing

\bibitem{agarwal_deep_2018}
\BIBentryALTinterwordspacing
B.~Agarwal, H.~Ramampiaro, H.~Langseth, and M.~Ruocco, ``A deep network model
  for paraphrase detection in short text messages,'' \emph{Information
  Processing \& Management}, vol.~54, no.~6, pp. 922--937, 2018. [Online].
  Available:
  \url{https://linkinghub.elsevier.com/retrieve/pii/S0306457317308713}
\BIBentrySTDinterwordspacing

\bibitem{tomar-etal-2017-neural}
\BIBentryALTinterwordspacing
G.~S. Tomar, T.~Duque, O.~T{\"a}ckstr{\"o}m, J.~Uszkoreit, and D.~Das, ``Neural
  paraphrase identification of questions with noisy pretraining,'' in
  \emph{Proceedings of the First Workshop on Subword and Character Level Models
  in {NLP}}.\hskip 1em plus 0.5em minus 0.4em\relax Copenhagen, Denmark:
  Association for Computational Linguistics, Sep. 2017, pp. 142--147. [Online].
  Available: \url{https://aclanthology.org/W17-4121}
\BIBentrySTDinterwordspacing

\bibitem{mikolov_linguistic_2013}
\BIBentryALTinterwordspacing
T.~Mikolov, W.-t. Yih, and G.~Zweig, ``Linguistic regularities in continuous
  space word representations,'' in \emph{Proceedings of the 2013 Conference of
  the North American Chapter of the Association for Computational Linguistics:
  Human Language Technologies}.\hskip 1em plus 0.5em minus 0.4em\relax
  Association for Computational Linguistics, 2013, pp. 746--751. [Online].
  Available: \url{https://aclanthology.org/N13-1090}
\BIBentrySTDinterwordspacing

\bibitem{mikolov2013efficient}
T.~Mikolov, K.~Chen, G.~Corrado, and J.~Dean, ``Efficient estimation of word
  representations in vector space,'' \emph{arXiv preprint arXiv:1301.3781},
  2013.

\bibitem{pennington2014glove}
\BIBentryALTinterwordspacing
J.~Pennington, R.~Socher, and C.~D. Manning, ``Glove: Global vectors for word
  representation,'' in \emph{Empirical Methods in Natural Language Processing
  (EMNLP)}, 2014, pp. 1532--1543. [Online]. Available:
  \url{http://www.aclweb.org/anthology/D14-1162}
\BIBentrySTDinterwordspacing

\bibitem{kalchbrenner-etal-2014-convolutional}
\BIBentryALTinterwordspacing
N.~Kalchbrenner, E.~Grefenstette, and P.~Blunsom, ``A convolutional neural
  network for modelling sentences,'' in \emph{Proceedings of the 52nd Annual
  Meeting of the Association for Computational Linguistics (Volume 1: Long
  Papers)}.\hskip 1em plus 0.5em minus 0.4em\relax Baltimore, Maryland:
  Association for Computational Linguistics, Jun. 2014, pp. 655--665. [Online].
  Available: \url{https://aclanthology.org/P14-1062}
\BIBentrySTDinterwordspacing

\bibitem{devlin_bert_2019}
\BIBentryALTinterwordspacing
J.~Devlin, M.-W. Chang, K.~Lee, and K.~Toutanova, ``{BERT}: Pre-training of
  deep bidirectional transformers for language understanding,'' in
  \emph{Proceedings of the 2019 Conference of the North American Chapter of the
  Association for Computational Linguistics: Human Language Technologies,
  Volume 1 (Long and Short Papers)}.\hskip 1em plus 0.5em minus 0.4em\relax
  Association for Computational Linguistics, 2019, pp. 4171--4186. [Online].
  Available: \url{https://aclanthology.org/N19-1423}
\BIBentrySTDinterwordspacing

\bibitem{lan2020albert}
Z.~Lan, M.~Chen, S.~Goodman, K.~Gimpel, P.~Sharma, and R.~Soricut, ``Albert: A
  lite bert for self-supervised learning of language representations,'' 2020.

\bibitem{raffel2020exploring}
C.~Raffel, N.~Shazeer, A.~Roberts, K.~Lee, S.~Narang, M.~Matena, Y.~Zhou,
  W.~Li, and P.~J. Liu, ``Exploring the limits of transfer learning with a
  unified text-to-text transformer,'' 2020.

\bibitem{clark2020electra}
K.~Clark, M.-T. Luong, Q.~V. Le, and C.~D. Manning, ``Electra: Pre-training
  text encoders as discriminators rather than generators,'' 2020.

\bibitem{dolan_automatically_2005}
\BIBentryALTinterwordspacing
W.~B. Dolan and C.~Brockett, ``Automatically constructing a corpus of
  sentential paraphrases,'' in \emph{Proceedings of the Third International
  Workshop on Paraphrasing ({IWP}2005)}, 2005. [Online]. Available:
  \url{https://aclanthology.org/I05-5002}
\BIBentrySTDinterwordspacing

\bibitem{dey-etal-2016-paraphrase}
\BIBentryALTinterwordspacing
K.~Dey, R.~Shrivastava, and S.~Kaushik, ``A paraphrase and semantic similarity
  detection system for user generated short-text content on microblogs,'' in
  \emph{Proceedings of {COLING} 2016, the 26th International Conference on
  Computational Linguistics: Technical Papers}.\hskip 1em plus 0.5em minus
  0.4em\relax Osaka, Japan: The COLING 2016 Organizing Committee, Dec. 2016,
  pp. 2880--2890. [Online]. Available: \url{https://aclanthology.org/C16-1271}
\BIBentrySTDinterwordspacing

\bibitem{attention-is-all-2017}
\BIBentryALTinterwordspacing
A.~Vaswani, N.~Shazeer, N.~Parmar, J.~Uszkoreit, L.~Jones, A.~N. Gomez,
  L.~Kaiser, and I.~Polosukhin, ``Attention is all you need,'' \emph{CoRR},
  vol. abs/1706.03762, 2017. [Online]. Available:
  \url{http://arxiv.org/abs/1706.03762}
\BIBentrySTDinterwordspacing

\bibitem{labelstudio}
\BIBentryALTinterwordspacing
``Labelstudio.'' [Online]. Available: \url{https://labelstud.io/}
\BIBentrySTDinterwordspacing

\bibitem{wieting-etal-2017-learning}
\BIBentryALTinterwordspacing
J.~Wieting, J.~Mallinson, and K.~Gimpel, ``Learning paraphrastic sentence
  embeddings from back-translated bitext,'' in \emph{Proceedings of the 2017
  Conference on Empirical Methods in Natural Language Processing}.\hskip 1em
  plus 0.5em minus 0.4em\relax Copenhagen, Denmark: Association for
  Computational Linguistics, Sep. 2017, pp. 274--285. [Online]. Available:
  \url{https://aclanthology.org/D17-1026}
\BIBentrySTDinterwordspacing

\bibitem{zhang_paws_2019}
\BIBentryALTinterwordspacing
Y.~Zhang, J.~Baldridge, and L.~He, ``{PAWS}: Paraphrase adversaries from word
  scrambling,'' in \emph{Proceedings of the 2019 Conference of the North
  American Chapter of the Association for Computational Linguistics: Human
  Language Technologies, Volume 1 (Long and Short Papers)}.\hskip 1em plus
  0.5em minus 0.4em\relax Association for Computational Linguistics, 2019, pp.
  1298--1308. [Online]. Available: \url{https://aclanthology.org/N19-1131}
\BIBentrySTDinterwordspacing

\bibitem{scherrer_tapaco_2020}
\BIBentryALTinterwordspacing
Y.~Scherrer, ``{TaPaCo}: A corpus of sentential paraphrases for 73 languages,''
  in \emph{Proceedings of the 12th Language Resources and Evaluation
  Conference}.\hskip 1em plus 0.5em minus 0.4em\relax European Language
  Resources Association, 2020, pp. 6868--6873. [Online]. Available:
  \url{https://aclanthology.org/2020.lrec-1.848}
\BIBentrySTDinterwordspacing

\bibitem{yang-etal-2015-wikiqa}
\BIBentryALTinterwordspacing
Y.~Yang, W.-t. Yih, and C.~Meek, ``{W}iki{QA}: A challenge dataset for
  open-domain question answering,'' in \emph{Proceedings of the 2015 Conference
  on Empirical Methods in Natural Language Processing}.\hskip 1em plus 0.5em
  minus 0.4em\relax Lisbon, Portugal: Association for Computational
  Linguistics, Sep. 2015, pp. 2013--2018. [Online]. Available:
  \url{https://aclanthology.org/D15-1237}
\BIBentrySTDinterwordspacing

\bibitem{guo2023close}
B.~Guo, X.~Zhang, Z.~Wang, M.~Jiang, J.~Nie, Y.~Ding, J.~Yue, and Y.~Wu, ``How
  close is chatgpt to human experts? comparison corpus, evaluation, and
  detection,'' 2023.

\bibitem{zhong-etal-2020-neural}
\BIBentryALTinterwordspacing
W.~Zhong, D.~Tang, Z.~Xu, R.~Wang, N.~Duan, M.~Zhou, J.~Wang, and J.~Yin,
  ``Neural deepfake detection with factual structure of text,'' in
  \emph{Proceedings of the 2020 Conference on Empirical Methods in Natural
  Language Processing (EMNLP)}.\hskip 1em plus 0.5em minus 0.4em\relax Online:
  Association for Computational Linguistics, Nov. 2020, pp. 2461--2470.
  [Online]. Available: \url{https://aclanthology.org/2020.emnlp-main.193}
\BIBentrySTDinterwordspacing

\bibitem{tang2023science}
R.~Tang, Y.-N. Chuang, and X.~Hu, ``The science of detecting llm-generated
  texts,'' 2023.

\bibitem{kirchenbauer2023reliability}
J.~Kirchenbauer, J.~Geiping, Y.~Wen, M.~Shu, K.~Saifullah, K.~Kong,
  K.~Fernando, A.~Saha, M.~Goldblum, and T.~Goldstein, ``On the reliability of
  watermarks for large language models,'' 2023.

\bibitem{song2020adversarial}
C.~Song, A.~M. Rush, and V.~Shmatikov, ``Adversarial semantic collisions,''
  2020.

\bibitem{zhang2019paws}
Y.~Zhang, J.~Baldridge, and L.~He, ``Paws: Paraphrase adversaries from word
  scrambling,'' 2019.

\bibitem{10.1145/3351095.3375623}
\BIBentryALTinterwordspacing
L.~Liang and D.~E. Acuna, ``Artificial mental phenomena: Psychophysics as a
  framework to detect perception biases in ai models,'' in \emph{Proceedings of
  the 2020 Conference on Fairness, Accountability, and Transparency}, ser. FAT*
  '20.\hskip 1em plus 0.5em minus 0.4em\relax New York, NY, USA: Association
  for Computing Machinery, 2020, p. 403–412. [Online]. Available:
  \url{https://doi.org/10.1145/3351095.3375623}
\BIBentrySTDinterwordspacing

\bibitem{aylin_sem_biases}
\BIBentryALTinterwordspacing
A.~Caliskan, J.~J. Bryson, and A.~Narayanan, ``Semantics derived automatically
  from language corpora contain human-like biases,'' \emph{Science}, vol. 356,
  no. 6334, pp. 183--186, 2017. [Online]. Available:
  \url{https://www.science.org/doi/abs/10.1126/science.aal4230}
\BIBentrySTDinterwordspacing

\bibitem{10.1145/3461702.3462536}
\BIBentryALTinterwordspacing
W.~Guo and A.~Caliskan, ``Detecting emergent intersectional biases:
  Contextualized word embeddings contain a distribution of human-like biases,''
  in \emph{Proceedings of the 2021 AAAI/ACM Conference on AI, Ethics, and
  Society}, ser. AIES '21.\hskip 1em plus 0.5em minus 0.4em\relax New York, NY,
  USA: Association for Computing Machinery, 2021, p. 122–133. [Online].
  Available: \url{https://doi.org/10.1145/3461702.3462536}
\BIBentrySTDinterwordspacing

\bibitem{parahrase_fake_news}
A.~Gautam and K.~R. Jerripothula, ``Sgg: Spinbot, grammarly and glove based
  fake news detection,'' in \emph{2020 IEEE Sixth International Conference on
  Multimedia Big Data (BigMM)}, 2020, pp. 174--182.

\bibitem{paraphrase_mis_2021}
A.~Huertas-Garc\'{\i}a, J.~Huertas-Tato, A.~Mart\'{\i}n, and D.~Camacho,
  ``Countering misinformation through semantic-aware multilingual models,'' in
  \emph{Intelligent Data Engineering and Automated Learning – IDEAL 2021:
  22nd International Conference, IDEAL 2021, Manchester, UK, November 25–27,
  2021, Proceedings}.\hskip 1em plus 0.5em minus 0.4em\relax Berlin,
  Heidelberg: Springer-Verlag, 2021, p. 312–323.

\end{thebibliography}
\newpage
\vspace{11pt}
\begin{IEEEbiographynophoto}{Chao Zhou}
As a second-year Master of Science student in the Computer Science department at Syracuse University, Chao Zhou was a Research Assistant in the Science of Science and Computational Discovery Lab under the supervision of Daniel Acuña. With a focus on paraphrase detection and sentence generation, Chao was working on providing a type-balanced paraphrase dataset to the plagiarism detection community. He was also interested in increasing the diversity of auto-generated sentences and documents using deep-learning models. He achieved his Master of Engineering degree from Syracuse University in May 2023 and his Bachelor of Engineering degree from the Chengdu University of Information and Technology in 2020. Chao Zhou is currently employed at an AI startup named SingularDance, serving as an NLP Engineer.
\end{IEEEbiographynophoto}
\begin{IEEEbiographynophoto}{Cheng Qiu}
As a foruth-year undergraduate at Vanderbilt University's School of Arts and Science, Cheng Qiu is at the intersection of applied mathematics and artificial intelligence, with research interests in natural language processing and computer vision. He was a research assistant in the Science of Science and Computational Discovery Lab under the supervision of Daniel Acuña as part of Syracuse University's SCORE grant for undergraduate students. He is currently working on projects involving biases in medical research and the application of medical imaging.
\end{IEEEbiographynophoto}
\begin{IEEEbiographynophoto}{Lizhen Liang}
A doctoral candidate at Syracuse University's School of Information Studies, majoring in information science and technology with a focus on computational social science. His research aims to detect and reduce inequality in science by utilizing advanced artificial intelligence techniques such as natural language processing, statistical analysis, and network science. He is particularly interested in the disparity between AI research conducted by industry-led teams and university-led teams, exploring how the dominance of large tech companies may affect resource-limited researchers in academia. Liang’s work is driven by a commitment to making AI beneficial for society, and his research investigates issues of AI safety, ethics, and social bias. He delivered a TEDx Talk in 2020 on identifying social biases in AI models, and plans to graduate in 2025.
\end{IEEEbiographynophoto}
\begin{IEEEbiographynophoto}{Daniel E. Acuna}
As the leader of the Science of Science and Computational Discovery Lab at the University of Colorado at Boulder, Associate Professor Daniel Acuna uses Machine Learning and A.I. to uncover the rules that drive successful knowledge production. His research in the field of computational social science focuses on understanding the historical relationships, mechanisms, and optimization opportunities of knowledge production. By harnessing vast datasets about publications and citations, he builds web-based software tools that accelerate knowledge discovery and has developed methods for detecting biases in A.I.

Recently, Daniel has been interested in the detection of scientific fraud and has created tools to improve literature search and peer review. His work has been funded by organizations such as NSF, DHHS, and the Sloan Foundation, and has been featured in outlets such as Nature News, NPR, and the Scientist. Through the SCORE project, he has also received funding from DARPA.
\end{IEEEbiographynophoto}    

\vfill
\end{document}